\documentclass[twoside,11pt]{article}

\usepackage[dvipsnames]{xcolor}\usepackage[dvipsnames]{xcolor}
\usepackage{subfigure}
\newcommand{\comment}[1]{}
\usepackage{mathtools}

\usepackage{jmlr2e}

\jmlrheading{X}{2021}{X-XX}{X/XX}{XX/XX}{Nat\'alia V. N. Rodrigues and L. Raul Abramo and Nina S. T. Hirata}

\ShortHeadings{Noise is Information}{N. Rodrigues, L.R. Abramo, N. Hirata}
\firstpageno{1}

\begin{document}

\title{
%Improving classification in 
%convolutional neural networks by including uncertainties \\
%or \\
%Accounting for attribute uncertainties in classification with %convolutional neural networks\\
%or \\
The information of attribute uncertainties: \\
what convolutional neural networks can learn about \\
errors in input data
}

\author{\name Nat\'alia V. N. Rodrigues \email               natalia.villa.rodrigues@usp.br \\
        \addr Instituto de F\'{\i}sica, Universidade de S\~ao Paulo, \\
        R. do Mat\~ao 1371, CEP 05508-090, S\~ao Paulo (SP), Brazil
        \AND
        \name L. Raul Abramo \email abramo@if.usp.br \\
        \addr Instituto de F\'{\i}sica, Universidade de S\~ao Paulo, \\
        R. do Mat\~ao 1371, CEP 05508-090, S\~ao Paulo (SP), Brazil
        \AND
        \name Nina S. Hirata \email nina@ime.usp.br \\
        \addr Instituto de Matem\'atica e Estat\'{\i}stica, Universidade de S\~ao Paulo, \\
        R. do Mat\~ao 1010, CEP 05508-090, S\~ao Paulo (SP), Brazil  \\}

\editor{} %Leslie Pack Kaelbling

\maketitle

\begin{abstract}%   <- trailing '%' for backward compatibility of .sty file
Errors in measurements are key to weighting the value of data, but are often neglected in Machine Learning (ML). 
We show how Convolutional Neural Networks (CNNs) are able to learn about the context and patterns of signal and noise, leading to improvements in the performance of classification methods.
We construct a model whereby two classes of objects follow an underlying Gaussian distribution, and where the features (the input data) have varying, but known, levels of noise.
This model mimics the nature of scientific data sets, where the noises arise as realizations of some random processes whose underlying distributions are known.
The classification of these objects can then be performed using standard statistical techniques (e.g., least-squares minimization or Markov-Chain Monte Carlo), as well as ML techniques. 
This allows us to take advantage of a maximum likelihood approach to object classification, and to measure the amount by which the ML methods are incorporating the information in the input data uncertainties.
We show that, when each data point is subject to different levels of noise (i.e., noises with different distribution functions), that information can be learned by the CNNs, raising the ML performance to at least the same level of the least-squares method -- and sometimes even surpassing it.
Furthermore, we show that, with varying noise levels, the confidence of the ML classifiers serves as a proxy for the underlying cumulative distribution function, but only if the information about specific input data uncertainties is provided to the CNNs.
\end{abstract}


\section{Introduction}

Machine Learning (ML) methods are becoming increasingly popular in the analysis of scientific data sets, especially in areas with large volumes of data such as high energy physics and astrophysics -- see, e.g., \citet{1992MNRAS.259P...8S,2003MNRAS.339.1195F,Baldi:2014kfa,MEHTA20191}. 
Typically, scientific data consists of individual measurements, each one with an associated uncertainty attached to it -- i.e., each input data point is assigned some probability distribution function (PDF) to represent the underlying noise distribution.
Traditional statistical techniques (e.g., Fisherian/frequentist or Bayesian methods) weigh the data according to their specific uncertainties in order to derive constraints and draw conclusions on the basis of those data sets \citep{Efron:1986,Tanabashi:2018oca}.
Nevertheless, despite the deep connections between machine learning and statistical inference \citep{MurphyBook}, measurement errors are routinely discarded in ML applications, even in the physical sciences \citep{Carleo:2019ptp}. 
In this paper we show the value of including the information content of the noise in Convolutional Neural Networks (CNNs), by quantifying the improvements in the performance of ML classifiers and comparing them against a baseline maximum likelihood method.

Scientific data sets can be highly complex to analyze, and it is not always clear what is the maximal amount of information that can be extracted from them, or how much of that information is being exploited by ML methods \citep{Chang:2017kvc}. As a result, it is often difficult to compare the performance of different methods or to assess improvements in the techniques.
In order to circumvent this difficulty, we created a toy model that is based on data with noise that is fully described by distributions that are known {\em a priori}, for each feature of the data set.
Moreover, the model parameters (which determine the class of the objects) are also drawn from Gaussian distributions, which means that we are able to quantify exactly the amount of mixing (confusion) between the classes.
These underlying distributions form the basis for our comparison between methods, allowing us to quantify the improvements we achieve by providing the uncertainties in the measurements to the CNNs.

The main issue we address in this paper is the value of the information contained in the noise associated with data (i.e., irreducible aleatoric uncertainty), and to what extent ML methods can learn about how to incorporate the information content about that noise. 
Clearly, if all the data points have exactly the same noise level, $x_i \to \bar{x}_i \pm \sigma$ (where $\bar{x}_i$ is the true value of that data point and $\sigma$ is the variance of the noise PDF), then there is zero additional information to be gained by explicitly providing that information to an algorithm. 
However, if different data points have different levels of uncertainties, $x_i \to \bar{x}_i \pm \sigma_i$, then each point contributes with a different weight to the determination of parameters.
The former case is known as homoscedastic errors, while the latter case corresponds to heteroscedastic errors -- see, e.g., \citet{10.5555/3295222.3295309}.
It is clear, from both a frequentist or a Bayesian viewpoint, that we should keep track of the different levels of signal and noise in data. 
The question is, then, what is the value of that information, and to what extent are the different ML techniques able to take it into account?

% Discutir um pouco sobre trabalhos anteriores que discutem inclusão de incertezas nas features.
The issue of attribute uncertainties in ML models has been addressed in different contexts
-- for recent reviews, see \citet{Abdar_2021} and \citet{Caldeira_2020}. Some early papers considered situations where heteroscedastic noise is a random variable which can be estimated nonparametrically using ML methods \citep{10.1145/1102351.1102413,Nix1994EstimatingTM}. For more recent applications using Random Forests, see \citet{Reis_2018}; for Support Vector Machines, see \citet{NIPS2004_22b1f2e0}; and for Neural Networks, see \citet{inproceedings}.

In this paper we quantify precisely how the output changes as the input data transitions from the homoscedastic (input data with uniform scatter) to the heteroscedastic (different, but known, scatters) regime.
We also use as a reference the maximum likelihood classification, which allows us not only to derive analytical formulas for the output errors, but also to find the exact uncertainty in the classification by running a Markov Chain Monte Carlo for each object in a test sample.
In particular, in this paper we show that CNNs are able to learn about the context of the information in the specific noise levels of the data, in such a way that they approach (and sometimes even surpass) the performance of the maximum likelihood approach.

%{\noindent \em Remainder omitted in this sample. See http://www.jmlr.org/papers/ for full paper.}

\section{The smiley-frowny model}\label{smiley-frowny}

Our toy model consists of two simple classes of objects: parabolic curves with positive and negative concavities. Hence, we have a binary classification problem where the positive and negative classes are convex (``smiley", $\smile$) and concave (``frowny", $\frown$), respectively. The basic idea is that each object is represented by a set of $n$ data points (or features), in such a way that each data point has an uncertainty that derives from some known probability density function (PDF). These uncertainties may be called the ``error bars'' of the measurements, which for our purposes can be thought of as the variances (second central momenta) of the PDFs. We discuss these uncertainties in detail in the next Section.

Since the underlying model (parabolic curves) is precisely known, we are able to classify each object using the method of maximum likelihood -- e.g., if we only require the class of the object, a least squares optimization method can be employed. Moreover, we are able to determine exactly the confidence of the classification of smiley and frowny objects using either a Fisher matrix approach or a Markov Chain Monte Carlo (MCMC) exploration of parameter space. The overall performance of the classification, as well as the confidence of the output for each object, can then be compared with ML models which do and do not include the input data uncertainties.

\begin{figure}
    \centering
    \includegraphics[width=1\linewidth]{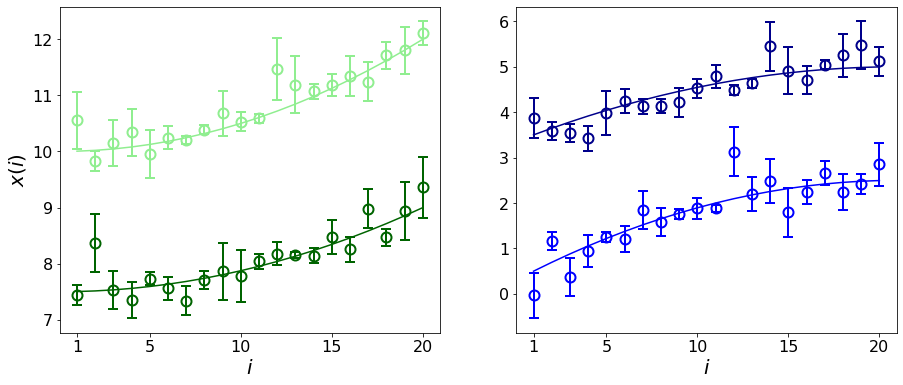}
    \caption{
    %\textit{Left}: PDFs from which the coefficients of the parabolic curves are sampled (see Table \ref{table:parameters}); \textit{Right}: 
    Parabolic curves generated according to the Smiley-Frowny model. The two lines in the left (green and dark green) are ``smiley'' objects, and the lines in the right (blue and dark blue) are ``frowny'' objects. 
    The vertical shifts (parameter $c$) of the curves shown were fixed for visualization purposes.
    The noisy measurements of the features of those objects are shown as data points with error bars.
    Each data point has a different, but known, PDF whose properties are summarized by the variances expressed in the error bars.
    }
    \label{fig:coeff pdf}
\end{figure}

It is important to stress that we are not trying to fit a curve. We are interested in employing a set of measurements in order to classify the objects.
%i.e., we will construct maps $X \to y$, where $X$ is a set of features, and $y = \{ 0, 1\} $ are the targets.
Our task is, therefore, to label curves as a smiley or a frowny in a binary classification scheme:
\begin{itemize}
    \item Positive class (1): $\smile$
    \item Negative class (0): $\frown$
\end{itemize}

The parabolic curves are generated according to the model:
\begin{eqnarray}
\label{Eq:xs}
\bar{x}_\smile (i) &=& a_\smile \, 
\left( \frac{i}{n} \right)^2 
+ b_\smile  \, \left( \frac{i}{n} \right) + c_\smile 
\\ \label{Eq:xf}
\bar{x}_\frown (i) &=& 
a_\frown  \, \left( \frac{i-n-1}{n} \right)^2 + b_\frown \, \left( \frac{i-n-1}{n} \right) + c_\frown \; , 
\end{eqnarray}
%$$ i_\frown = i_\smile - (\text{max}(i) + \text{min}(i)) $$
where the indices $i=1,2, \ldots, n$, with $n$ being the number of measurements, which are the attributes, or features, that characterize the curves.
The parameters $a,\ b$ and $c$ are drawn from normal distributions (see Fig. \ref{fig:coeff pdf}) with means and standard deviations specified in Table 
\ref{table:parameters}. The random nature of the parameters ensures that we have a variety of objects in each class, and the two PDFs for the curvature parameter $a$ are sufficiently separated (4 $\sigma$'s) that the probability that an object sampled from the distribution of one class has the sign of the other class is $3.17 \times 10^{-5}$, which is irrelevant for the purposes of our discussion.

With the definitions of Eqs. (\ref{Eq:xs}-\ref{Eq:xf}), both curves grow by the same amount from start to end: $\bar{x}(i=n)-\bar{x}(i=1) = b (n-1)/n  \, \pm \,  a \,   (n^2-1)/n^2$, where the plus and minus signs refer to the positive (convex, smiley) and negative (concave, frowny) classes, respectively. Since the distributions of the curvatures are anti-symmetric, $a_\smile \leftrightarrow - a_\frown$, all curves on average rise by the same amount from $i=1$ to $i=n$, which further mixes the two classes.
This feature of the model ensures that the only significant distinction between smiley and frowny objects is the concavity of the curves. Otherwise, a ML classifier could employ other patterns (such as whether the function grows or falls with $i$) to distinguish between the two classes and, thus, outperform a model-based maximum likelihood method such as least squares.

\begin{table}[!htbp]
\caption{Parameters of the normal distributions from which the coefficients of the parabolic curves are sampled.}     \label{table:parameters}
\centering
\begin{tabular}{l c c c c c}
\hline
\hline
Coefficient & $a_\smile$  & $a_\frown$ & $b$ & $c$ \\    
\hline
   $\mu$  & $1$ & -1 & 0 & 10 \\
   $\sigma$  & 0.25 & 0.25 & 1 & 3 \\
\hline
\end{tabular}
\end{table}

\section{Noise and information in input data: a toy model}
\label{toy_model}

Scientific data sets are comprised of measurements which are performed with the help of instruments with some nominal uncertainties. However, those uncertainties are usually not fixed for all time: even with the same instrument, some measurements may have higher or lower uncertainties depending on several conditions. Any experiment will carefully assess what those uncertainties are for each data point, taking into account the different circumstances under which those measurements were made -- see, e.g., \citet{TaylorBook}.

To be clearer, one can think of two main sources of aleatoric uncertainties. The first is the quality, or nominal sensitivity, of the apparatus used to perform the measurements: e.g., the lengths of nails that are measured with a micrometer are intrinsically more accurate than the same measurements made using a ruler. 
The second source arises from the different conditions under which the measurements are made by the same apparatus. As an example, one can think of the measurements of the lengths of those same nails with a micrometer, but in some days the temperature of the laboratory is more stable than others, resulting in different dilation factors for the nails. Another example comes from astronomical data sets: in that case, the nominal uncertainty is determined by the size of the telescope's mirror and the sensitivity of the detectors, among other factors. However, some nights are brighter than others, some objects appear close to bright sources of light, and so on and so forth, meaning that different images, as well as different parts of the same image, have varying degrees of data uncertainty.
Any careful experimenter will label which measurements have higher or lower levels of uncertainty as a result of those different conditions. 

In this work we construct a simple model to reproduce these varying degrees of uncertainty in scientific data. First, we assume that the nominal accuracy of the measuring instrument is given in terms of a parameter $\sigma_0$, meaning that under some ``ideal'' conditions for that instrument, the measurements are random numbers that follow a normal distribution with variance $\sigma_0$. 
And second, we introduce parameters $g_i$ that follow a uniform distribution, in such a way that the actual measurements $x_i$ (now under varying conditions) have uncertainties given by $g_i \sigma_0$. 
In other words, we have:
\begin{equation}
    \label{Eq:xdx}
x_i = \bar{x}_i + \delta x_i \; ,
\end{equation}
where $\bar{x}_i$ are the true values of the measurements and $\delta x_i$ are random numbers sampled from a Gaussian probability distribution functions with zero mean and variance $g_i \sigma_0$, i.e.:
\begin{align}
\label{Eq:dist}
p(\delta x_i) = 
\frac{1}{\sqrt{2\pi \, g_i^2 \, \sigma_0^2}} \, e^{-\frac12 \frac{\delta x_i^2}{g_i^2 \sigma_0^2} } \; .
\end{align}
Here $g_i$ are numbers that are {\em known} for each individual measurement: one can think of a label for each data point indicating the degree to which the uncertainties are higher or lower than the nominal ones. In order to simulate the varying conditions under which those measurements are performed, we draw the factors $g_i$ from a Uniform distribution in the interval:
$$
g_i \in \left[ \bar{g} -\Delta g, \bar{g} + \Delta g \right] \, ,
$$
where $\Delta g $ is the noise dispersion parameter, and the expectation value (mean) of that parameter given the Uniform distribution is $\langle g_i \rangle_U = \int dg \, g \, U(g) = \bar{g}$, where $U(g) = 1/(2\Delta g)$ for $ \bar{g} -\Delta g \leq g \leq \bar{g} + \Delta g$, and 
$U(g) = 0$ if $g < \bar{g} -\Delta g $ or
$g > \bar{g} + \Delta g $
(clearly, $0 \leq \Delta g < \bar{g}$).
For $\Delta g=0$ we have a homoscedastic dataset, and as this parameter grows, the degree of heterodasticity increases.
However, we stress the fact that the factors $g_i$ are {\it known} and, as opposed to the Gaussian random process underlying the noise, the values $g_i$ should not be regarded as stochastic variables in a fundamental sense: they are part of the information of the data set, and can be passed on to the ML methods.

Since the two distributions are uncorrelated by construction, the mean variance of the data errors can be easily computed:
\begin{align}
    \langle \delta x_i^2 \rangle_{U,G} 
    = \langle g_i^2 \rangle_U \, \sigma_0^2
    = \left( 1 + \frac13 \frac{\Delta g^2}{\bar{g}^2} \right) \, \bar\sigma_0^2  \; ,
\end{align}
where $\bar\sigma_0 = \bar{g} \, \sigma_0$ denotes the mean nominal noise.
This simple result tells us that when there are varying levels of noise in data, as described by this model, then the mean noise of the ensemble is actually {\em higher} than the mean nominal noise $\bar\sigma_0$.

Traditional statistical tools for data analysis are naturally equipped to deal with different levels of noise in input data. In particular, the likelihood is given by:
\begin{align}
\label{Eq:lik}
L = N \, \exp{ \left[ -\frac12 \sum_i^n \left( \frac{x_i - \bar{x}_i}{g_i \, \sigma_0} \right)^2 \right] } \; ,   \end{align}
where $N$ is some normalization and $\bar{x}_i$ is the expectation value of the variable $x_i$
(or, in this context, the ``theory'' that we would like to fit to the data).
In scientific applications we usually assume the theory to depend on a set of parameters denoted by the vector $\theta^\mu$ through some model, $\bar{x}_i(\theta^\mu)$ -- in our example, those parameters are $\theta^\mu = \{ a , b, c \}$, so $\mu = 1,2,3$.

The likelihood function tells us which regions in parameter space are preferred, given the model, the data, and the uncertainties (or, more generically, the data covariance). 
Although a more thorough exploration of the likelihood function in parameter space is usually carried out using Markov Chains generated via a Monte Carlo algorithm (in that respect, see Section \ref{results gauss}), we can estimate the shape of the likelihood using a Gaussian approximation, in which case the logarithm of the likelihood is a quadratic function.
The curvature of that multivariate quadratic function at the peak (maximum likelihood) is the Fisher information matrix, which is computed by means of the Hessian:
\begin{eqnarray}
    \label{Eq:Fish0}
    F[\theta^{\mu},\theta^{\nu}] 
    = - \left\langle \frac{\partial^2 \log L}{\partial \theta^\mu \, \partial \theta^\nu}
    \right\rangle 
    \; .
\end{eqnarray}
The inverse of the Fisher matrix yields an estimate of the parameter covariance, ${\rm Cov} [\theta^\mu,\theta^\nu] \to \{ F[\theta^{\mu},\theta^{\nu}] \}^{-1}$ -- see, e.g., \citet{Tanabashi:2018oca} for many examples and applications in the physical sciences.

For the likelihood function of Eq. (\ref{Eq:lik}) we obtain:
\begin{eqnarray}
    \label{Eq:Fish1}
    F[\theta^{\mu},\theta^{\nu}] 
    = 
    \sum_i^n \frac{\partial_\mu \, \bar{x}_i 
    \, \partial_\nu \bar{x}_i}{g_i^2 \, \sigma_0^2}
    \; ,
\end{eqnarray}
where $\partial_\mu (\cdots) = \partial (\cdots) /\partial \theta^\mu$, and we used the fact that the data is both unbiased, $\langle x_i \rangle = \bar{x}_i$, and that the ``measurements'' do not depend on the parameters, $\partial_\mu x_i = 0$ (the {\em theory}, on the other hand, obviously does: $\partial_\mu \bar{x}_i \neq 0$.) 

At this point we can take the expectation value over the uniform distribution of the noise dispersion to obtain the mean Fisher matrix:
\begin{eqnarray}
    \label{Eq:FishAv}
    \bar{F}[\theta^{\mu},\theta^{\nu}] 
    = \sum_i^n 
    \left\langle \frac{1}{g_i^2}
    \right\rangle_U
    \frac{\partial_\mu \bar{x}_i \, 
    \partial_\nu \bar{x}_i}{\sigma_0^2}
    =     \frac{1}{1-(\Delta g/\bar{g})^2}
    \sum_i^n 
    \frac{\partial_\mu \bar{x}_i \,
    \partial_\nu \bar{x}_i}{\bar\sigma_0^2}
    \; .
\end{eqnarray}
We recognize the sum on the right-hand-side as the Fisher matrix for a nominal uncertainty $\bar\sigma_0$. Therefore, the amount by which different data points may have different uncertainties (expressed by the noise dispersion parameter $\Delta g$) has the effect of {\em increasing} the Fisher matrix with respect to the case where the noise is fixed to the nominal value.

In other words: when each data point has a different (but known) uncertainty, even though the mean noise level is {\em higher}, the Fisher information actually {\em increases}. This happens, of course, because the specific noise levels in the data are acting as weights: noisier data are down-weighted, and less noisy data are up-weighted, resulting in higher discriminatory power. Here we are simply restating the fact that we always lose information if we do not distinguish between low-noise and high-noise measurements. This statement remains true when applied to ML techniques.

Our toy model for the smiley/frowny objects has a very simple, analytical Fisher matrix. Using Eq. (\ref{Eq:xs}) into the Fisher matrix of Eq. (\ref{Eq:FishAv}) we obtain: 
\begin{equation}
    \bar{F}[\{a,b,c\}] = \frac{1}{ [1-(\Delta g/\bar{g})^2] \, \bar\sigma_0^2}
    \sum_{i=1}^n
    \left(
    \begin{array}{ccc}
    (i/n)^4 & \pm  \, (i/n)^3 &  (i/n)^2  \\
    \pm \, (i/n)^3 & (i/n)^2 & \pm \, (i/n)^1  \\
    (i/n)^2 & \pm \, (i/n)^1 &  (i/n)^0
    \end{array}
    \right) \; ,
\end{equation}
where the plus and minus signs correspond to the smiley and frowny objects, respectively.
All terms in this matrix have a closed form, given by the sum rules:
\begin{eqnarray}
    \sum_i^n i^0 &=& n \\
    \sum_i^n i^1 &=& \frac{n(n+1)}{2} \\
    \sum_i^n i^2 &=& \frac{n(2n^2+3n+1)}{6} \\
    \sum_i^n i^3 &=& \frac{n^2(n^2+2n+1)}{4} \\
    \sum_i^n i^4 &=& \frac{n(n^4+15n^3+10 n^2-1)}{30} 
\end{eqnarray}

From these expressions we can compute the mean uncertainty in the parameter $a$ -- which is exactly the same for the two classes, since they are symmetric in the sense that $a_\smile \leftrightarrow - a_\frown$. Upon inverting the Fisher matrix we obtain the covariance matrix, whose diagonal term corresponding to the parameter $a$ yields the result: 
\begin{align}
    \label{Eq:Cova}
    \Sigma_a^2 = \left( \bar{F}^{-1} \right)_{aa} = 
    \bar\sigma_0^2 \,
    \left( 1-\frac{\Delta g^2}{\bar{g}^2} \right) \,  
    \frac{180 \, n^3}{n^4 - 5 n^2 + 4} \; ,
\end{align}
where we defined the posterior uncertainty of the parameter $a$ by $\Sigma_a$, which should not be confused with the width of the parent distribution for that parameter, $\sigma_a$, that expresses the intrinsic (true) diversity of objects in our classes.
As discussed above, the uncertainty in the class of the objects is lower when the input data has varying levels of noise. 
Furthermore, with only two points ($n=2$) the class is completely undetermined -- as it should, since with two points it is impossible to derive the curvature. 
For larger values of $n$ the uncertainty scales as $\Sigma_a \sim 1/\sqrt{n}$.

We can further define a mean confidence for a maximum likelihood classification. The variance obtained above can be used in a normal distribution for the parameter $a$, and integrated for positive and/or negative values, yielding the probability that an object is either in the positive or negative class. E.g., the probability that an object has $a>0$ is given by:
\begin{eqnarray}
\nonumber
    P(a>0) &=& 
    \int_{-\infty}^\infty da 
    \frac{1}{\sqrt{2\pi \, \sigma_a^2}} 
    \, e^{-\frac12 \frac{(a-\mu_a)^2}{\sigma_a^2}}
    \int_0^\infty da' 
    \frac{1}{\sqrt{2\pi \, \Sigma_a^2}} 
    \, e^{-\frac12 \frac{(a'-a)^2}{\Sigma_a^2}}
    \\ \label{Eq:err_a}
    &=& \frac12 + \frac12 {\rm Erf} \left[ \frac{\mu_a}{\sqrt{2(\sigma_a^2 + \Sigma_a^2)}} \right] \, ,
\end{eqnarray}
where ${\rm Erf} (z) = 2/\sqrt{\pi} \int_0^z dt \, e^{-t^2}$ is the error function, and $\mu_a$ and $\sigma_a$ are, respectively, the central value and variance of the distribution for the parameter $a$ -- see Table \ref{table:parameters}. The expression above is therefore the average confidence of the maximum likelihood classification -- and, of course, it also expresses the cumulative distribution function at the value $\mu_a$.

\begin{figure}
    \centering
      \includegraphics[width=0.6\linewidth]{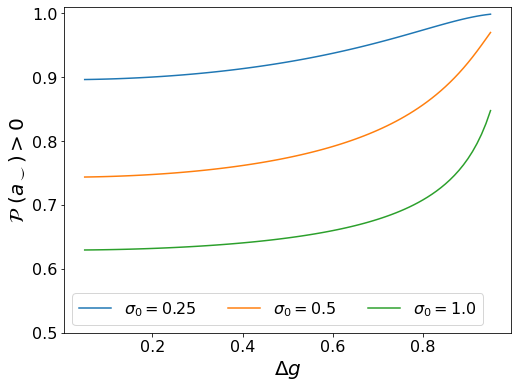}
      %\label{fig:su  %\caption{A subfigure}
    \caption{Mean probability that smiley objects are correctly classified.
    From top to bottom, the curves correspond to the parameters $\sigma_0 = 0.25$ (blue line), 0.5 (orange) and 1.0 (green), respectively, and we used $n=20$ features.
    }
    \label{fig:Prob_a}
\end{figure}

In Fig. \ref{fig:Prob_a} we plot the probability of Eq. (\ref{Eq:err_a}) for $\mu_a=1$ (smiley class), as a function of $\Delta g$. 
In this example we used $\bar{g}=1$ and each object has $n=20$ features (data points). 
From top to bottom, the curves correspond to increasing values of the nominal input error, $\bar\sigma_0 = \sigma_0 = 0.25$, 0.5, and 1.0.
For very high uncertainties in the input data ($\sigma_0 \gg 1.0$) the probability approaches 0.5, which means zero confidence in the classification. That confidence grows as we lower the nominal uncertainty and/or if we increase the noise dispersion parameter $\Delta g$.
\comment{
\begin{figure}
    \centering
      \includegraphics[width=0.5\linewidth]{smiley_frowny/Prob_asmile_pos.png}
      %\label{fig:su  %\caption{A subfigure}
    \caption{Mean probability that smiley objects are correctly classified.
    From top to bottom, the curves correspond to the parameters $\sigma_0 = 0.25$ (blue line), 0.5 (orange), 1.0 (green) and 2.0 (red line), respectively, and we used $n=20$ features.
    }
    \label{fig:Prob_a old}
\end{figure}
}

However, as much as the calculations above are able to provide insights into the problem at hand, we would be misguided if we attempted to use Eq. (\ref{Eq:err_a}) to infer the confidence for a maximum likelihood classification of individual objects, for two reasons. 
First, the Cramér-Rao-Fréchet bound [see, e.g, \citet{Efron:1986}] implies that the Fisher estimator has minimal variance, meaning that the probability expressed by Eq. (\ref{Eq:err_a}) is an extreme, limiting case.
The second reason is that we approximated the actual Fisher matrix, Eq. (\ref{Eq:Fish1}), by an average over the (uniformly distributed) noise dispersion parameter, which resulted in Eq. (\ref{Eq:FishAv}). 
Due to the random nature of the specific uncertainty $g_i$ in our model, individual objects may have better (less noisy) or worse (noisier) data points, and for those objects the confidence will differ from what is expressed by Eq. (\ref{Eq:err_a}).
Hence, Eq. (\ref{Eq:err_a}) represents an ideal scenario: in practice, applying the least squares method for a sample of objects results in an average accuracy which is slightly worse than the one that results from using this analytical formula.
Therefore, the exploration of the likelihood in parameter space shall be performed object-by-object according to Eq. (\ref{Eq:lik}), using either a least squares method (if one is only interested in the class itself) or an MCMC (if we also need to know the probability of the classification). These maximum likelihood results then form the basis for our comparison with the performance and confidence of the classification using ML methods.

It is instructive, in the context of this Section, to also consider an associated linear problem. 
A linear estimator for the curvature is given by:
\begin{equation}
    \label{Eq:LinProb1}
    \hat{a}_\mu = n^2 \sum_{i=1}^n M_{\mu i} \, x_i \; ,
\end{equation}
where $M_{\mu i} = \frac12 \delta_{\mu ,j+1} + \frac12 \delta_{\mu ,j-1} - \delta_{\mu ,j} $, and the indices $\mu =2,3, \ldots, n-1$ can be regarded as the $n-2$ intermediate points where we are able to estimate the curvature ($a$) through differences of the neighboring points. 
It is easy to check that the linear estimator is unbiased, and independent of the model parameters $b$ and $c$: just use the identities $\sum_i M_{\mu i} = \sum_i M_{\mu i} \, i = 0$, and  $\sum_i M_{\mu i}\, i^2 = 1$. 

The covariance of the linear estimator is given by the expectation value:
\begin{equation}
    \label{Eq:LinProb2}
    C_{\mu\nu} =
    \langle (\hat{a}_\mu -a)( \hat{a}_\nu - a) \rangle = n^4 \sum_{i=1}^n M_{\mu i} \, M_{\nu i} \,  g_i^2 \, \sigma_0^2 
    \; ,
\end{equation}
which, after averaging over the uniform distribution for the noise dispersion leads to:
\begin{equation}
    \bar{C}_{\mu\nu} =
    \left( 1 + \frac13 \frac{\Delta g^2}{\bar{g}^2} \right) \, \bar\sigma_0^2 \, n^4 \sum_{i=1}^n M_{\mu i} \, M_{\nu i} \; .
\end{equation}
It is straightforward (though by no means trivial) to check that the mean Fisher information matrix corresponding to this linear estimator is given by: 
\begin{align}
    \label{Eq:Covay}
    \Sigma_{a,lin}^2 
    = \left[ \sum_{\mu\nu} \bar{C}_{\mu\nu}^{-1} \right]^{-1} = 
    \bar\sigma_0^2 \,
    \left( 1+\frac13 \frac{\Delta g^2}{\bar{g}^2} \right) \,  
    \frac{180 \, n^3}{n^4 - 5 n^2 + 4} \; ,
\end{align}
which can be compared with Eq. (\ref{Eq:Cova}). 

Equation (\ref{Eq:Covay}) highlights a property that was already apparent in Eq. (\ref{Eq:Cova}), which is the fact that the uncertainty in the output (the curvature, $a$) depends on the number of features only through the factor $180 \, n^3/(n^4 - 5n^2 + 4)$. 
The output uncertainty on the pattern of input errors, on the other hand, is encapsulated by its dependence on $\Delta g$. 
When we include the information about specific noise levels through inverse covariance weighting, as expressed by Eq. (\ref{Eq:Fish1}), that pre-factor is $1-\Delta g^2/\bar{g}^2$ -- i.e., providing the information about noise {\em improves} the constraints.
However, when we neglect the error information and resort to direct estimators such as $\hat{a}_\mu$, then that pre-factor becomes $1+\Delta g^2/(3\bar{g}^2)$, {\em increasing} the output uncertainties.

Finally, it is worth pointing out the limitations of tools such as Tikhonov regularization, which are often used to prevent overfitting and to minimize empirical error -- see, e.g., \citet{StatLearn}, as well as related methods such as the one proposed by \citet{inproceedings}.
Basically, regularization techniques work by effectively imposing a threshold on very small eigenvalues in ill-posed inverse linear problems. 
However, the associated linear problem presented above is perfectly well-posed, and still it results in a degradation of the output uncertainties when compared with the optimal (inverse covariance weighting) estimator. 
Furthermore, the covariance of the linear estimator, Eq. (\ref{Eq:LinProb2}), is a positive-definite matrix, $\sum_{\mu\nu} C_{\mu\nu} V_\mu V_\nu \geq 0$ for any Real-valued vector $V_\mu$, so all the eigenvalues of this covariance matrix are real, non-negative numbers.
Consequently, imposing any kind of minimum threshold for those eigenvalues would in fact {\em increase} the linear estimator uncertainty, Eq. (\ref{Eq:Covay}), which means that no amount of regularization can possibly compensate the lack of information about the noise of each input data point.

\comment{
\subsection{Least Squares and MCMC: the baseline classifiers}\label{ml classifiers}
%\textcolor{red}{ideia: mostrar como o least squares varia com diferentes curvas. Assim dá para ter uma ideia da "variabilidade" da performance com os dados em si, independente dos modelos de ML.}
The baseline for curve fitting is the least squares optimizer, which account for the uncertainties of the measured points in the curve.
We know precisely the analytic expression for the smiley-frowny curves, which means that we can use a least-squares optimizer to find the optimal set parameters $a, b, c$ to fit the parabolic curves. However, we are only interested in the value of the parameter $a$, which characterizes a curve as smiley or frowny. The least-squares optimizer can be interpreted as a classifier based on the value of a, which will be greater or lower than zero.
}

\section{Machine Learning Classifiers}

Our problem statement is to classify sets of features that characterize curves which belong to a given class. We can model this as a supervised machine learning classification task. Our training sample is the set of tuples $\{ \mathbf{x}^{(j)}, y^{(j)} \}_{j = 1}^{m}$, where $\mathbf{x}^{(j)} \in \mathbb{R}^n$ are the smiley-frowny parabolic curves and $y^{(j)} \in \{0, 1\}$ are the corresponding labels. Additionally, we have the uncertainties $\sigma^{(j)} \in \mathbb{R}^{n}$ for each $\mathbf{x}^{(j)}$, for those models that use that information.
\comment{
{\color{purple}[Por que precisamos do índice $i$ ?]}
{\color{Blue}[Os índices $i$ representam o vetor de dados para cada objeto, $i=1,2,\ldots,n$. Ou seja, cada dado tem $n$ atributos. A única dúvida é se esses $n$ incluem os erros ou não, e como expressar isso nessa notação. Natália?...]}
{\color{orange}[os n não incluem os erros, nessa notação as features são apenas as medidas.]}
{\color{orange}[sugestao: Our problem statement is to classify sets of features that characterize curves which belong to a given class. We can model this as a supervised machine learning classification task. Our training set is the set of tuples $\{ \mathbf{x}^{(j)}, y^{(j)} \}_{j = 1}^{m}$, where $\mathbf{x}^{(j)} \in \mathbb{R}^{2n}$ is the set of $n$ points of the $j^\text{th}$ smiley-frowny parabolic curve (see Eq. \eqref{Eq:xdx}) followed by the $n$ corresponding uncertainties, and $y^{(j)} \in \{0, 1\}$ is the label. Isto é, $\mathbf{x}^{(j)} = [x^{(j)}_i, \sigma^{(j)}_i],\ i = [1, 2, \dots, n]$]}
{\color{purple}Ok, entendi que o $i$ não deve estar em $\{ \mathbf{x}^{(j)}_i, y^{(j)} \}_{j = 1}^{m}$. Sobre  $\mathbf{x}^{(j)} = [x^{(j)}_i, \sigma^{(j)}_i] \in  \mathbb{R}^{2n}$, não sei se seria a melhor solução. Talvez possamos dizer que os dados (sample) são $\{ \mathbf{x}^{(j)}, y^{(j)} \}_{j = 1}^{m}$ e que adicionalmente temos também $\sigma^{(j)} \in \mathbb{R}^{n}$ para cada $\mathbf{x}^{(j)}$ ?}
}

Due to the nature of the data, where the features are sorted in a significant way, we find it more appropriate to use CNNs since they are able to recognize local patterns. Multiple problems of sequential data analysis are in fact tackled with CNNs and 1D convolutional kernels -- see, e.g., 
\citet{article, busca2018quasarnet, Cabayol_2018, Ismail_Fawaz_2019, mozaffari2020review, rs13081519}.
In this work we implemented the classifiers with \texttt{keras} \citep{chollet2015keras}.

To classify the smiley-frowny curves, we start with a baseline dataset with the parameters described in Table \ref{table:baseline sf} and in Section \ref{results gauss} we explore multiple scenarios where we deviate from this baseline dataset by varying the values of the parameters. Unless noted otherwise, our networks are trained with a sample of $m = 2 \times 10^5$, from which $80\%$ are taken to train and the remaining $20\%$ are used to validate the models, i.e., to fix the hyperparameters of the networks and to monitor the bias-variance trade-off. The results shown in the next sections were computed with test sets, which were left completely unbiased by the training procedure and contains $m = 10^5$ instances. All the data sets are balanced in terms of classes, i.e., they contain the same amount of smiley and frowny objects.

\begin{table}[!htbp]
\caption{Baseline smiley-frowny dataset parameters.} \label{table:baseline sf}
\centering
\begin{tabular}{c c c c c c}
\hline
\hline
$m$ (training set) & $n$ & $\sigma_0$ & $\bar g$ & $\Delta g$ \\
\hline
   $2 \times 10^5$  & $20$ & $0.5$ & $0.6$ & $0.5$\\
\hline
\end{tabular}
\end{table}

We created multiple CNNs which differ from each other mainly in the content and shape of the input data. 
The specifications of each version are described in the next subsections. The general training settings are the following. We used the binary cross-entropy loss function
and the \texttt{Adam} \citep{kingma2017adam} optimizer. In all intermediate layers we used the ReLU activation function and in the last layer we used the Softmax activation function so that the scores of both classes sum up to one and, thus, we have a probabilistic interpretation for the output. The convergence of the training was monitored with learning curves for accuracy and for the loss function at each iteration (epoch) for both training and validation sets and we used batches of size $100$ to train the networks.
We used the \texttt{EarlyStopping} callback conditioned to the validation set loss score with \textit{patience} of 16 epochs and the \texttt{ReduceLROnPlateau} callback to reduce the learning rate when the validation set loss stagnate for 10 epochs.
The final set of weights is the one corresponding to the epoch with best accuracy in the validation set.

%For our reference set of parameters, namely, Gauss noise, $n=20$, $m=200k$ we created a ``standard" architecture with three convolution layers, with (32, 64, 64) kernels respectively, and 2 dense layers, the intermediate with 64 and the output with 2 neurons. After each convolution, we add a BatchNormalization and a MaxPooling layer. The shape of the kernels depend on the input data and vary for each model (see below). The architectures might deviate from our ``standard", depending, mainly on the degree of the uncertainty. As the curves become noisier, we noticed that more regularization was required for the learning curves to show a good behaviour. In different scenarios, mainly depending on the degree of the noise, the architectures might deviate from our ``standard" one.

\comment{
\begin{figure*}
        \centering
        \begin{subfigure}
        \centering
            \includegraphics[width=.5\linewidth]{classifiers/cnn1D_input.png}
            %\label{fig:su  %\caption{A subfigure}
        \end{subfigure}%
        \\ \vskip 0.5cm
        \begin{subfigure}%{0.5\textwidth}
        \centering
            \includegraphics[width=.8\linewidth]{classifiers/gauss_images.png}
            %\caption{A subfigure}
            %\label{fig:sub2}
        \end{subfigure}
    \caption{Input data illustration. In the top we present the input formats for the three CNN1D networks, $x_i$ are the measurements and $\sigma_i = g_i \sigma_0$ are the uncertainties of each measurement. 
    On the bottom panels we show how to convert the input data (left) to a matrix, or image (right) for the CNN2D network. 
    The black rectangle represents one of the convolution kernels.}
    \label{fig:input data}
    \end{figure*}
}

\begin{figure*}
    \centering
    \includegraphics[width=1.05\linewidth]{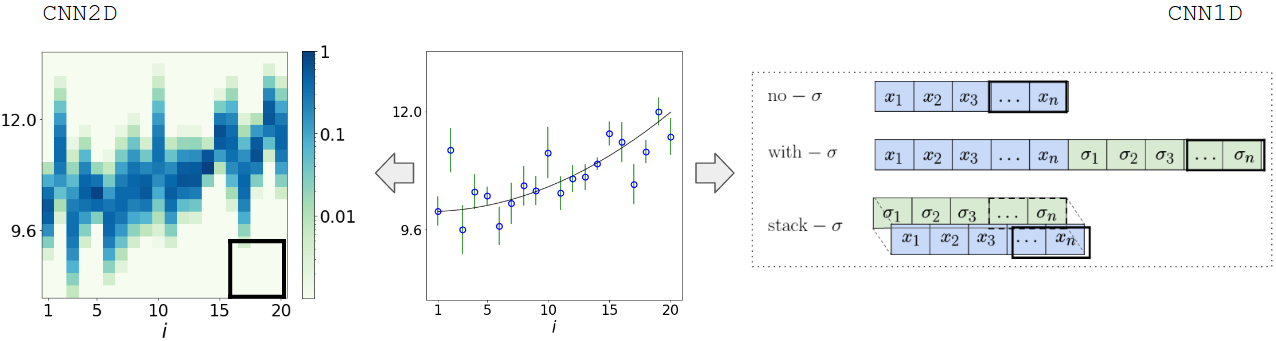}
    \caption{Input data illustration. On the right we present the input formats for the three CNN1D networks, $x_i$ are the measurements and $\sigma_i = g_i \sigma_0$ are the uncertainties of each measurement. 
    On the left we show how to convert the input data (center) to a matrix, or image, for the CNN2D network (notice that this is an image with a single channel, where the color gradient indicates the values of each pixel). 
    The black rectangles represent the convolution kernels.}
    \label{fig:input data}
    \end{figure*}

We now turn to the description of the different networks that we trained in order to classify the {\em smiley} and {\em frowny} objects.

\subsection{CNN1D}

In the CNN1D models, the input data shapes are 1D vectors, and, thus, the convolution kernels are also 1D. We compared three versions of CNN1D models (see Fig. \ref{fig:input data}):

\begin{itemize}
    \item \textbf{no-$\sigma$}: the input data shape is $(n,\ 1)$, it contains only the $n$ measurements.
\end{itemize}

\begin{itemize}
    \item \textbf{with-$\sigma$:} the input data shape is $(2 \cdot n,\ 1)$ where the $n$ measurements are followed by the $n$ corresponding uncertainties. This is a first approach to include uncertainties, but without making any hypothesis on what is the best way to represent this additional information.
\end{itemize}

\begin{itemize}
    \item \textbf{stack-$\sigma$}: the input data is the set of measurements and errors arranged in channels. The input shape, therefore, is $(n,\ 2)$. It is identical to CNN1D with-$\sigma$ in terms of the available information, but, in this case, the spatial relation between the measurements and corresponding errors is represented in a straightforward way in this model, providing a context for the uncertainties.
    %Obviously, this setup is completely equivalent to organizing the features as 2D matrices with only 2 rows -- one for the measurements and one for errors. But then, one would need to work with 2D kernels.
\end{itemize}
For the baseline training set (Table \ref{table:baseline sf}), we defined a simple standard network for all three CNN1D versions, which consists of three convolution layers with kernel shapes  $(5,),\ (3,)$ and $(3,)$ with $32, 64, 64$ filters, respectively, one intermediate dense layer with 64 neurons and, finally, the output layer with $2$ neurons. Each convolution layer does \textit{padding}, i.e, the output feature map has the same size as the input feature map, and is followed by a \texttt{MaxPooling} layer with kernel size and stride $(2,)$. We also add \texttt{BatchNormalization} and dropout layers with dropout rates typically between $0.2$ and $0.4$. As we deviate from the baseline dataset, some modifications on this standard architecture might be necessary to ensure the convergence of the models. In particular, as we increase the level of noise in the data by either decreasing the number of features $n$ or increasing the nominal error $\sigma_0$, a less complex network (with a lower number of layers) or a more regularized network is more appropriate because, as the data becomes noisier, the models are more likely to overfit \citep{10.5555/2207825}.% [ref, discussion]

\subsection{CNN2D images}\label{images section}

Several problems in the physical sciences have benefited from the power of ML methods that were developed for the analysis of images, in particular applications developed for high energy physics \citep{Baldi:2014kfa} or astrophysics \citep{2007ApJ...660.1176E}. 
In the CNN2D method we build on the same idea, however we use the additional dimension to represent the uncertainties in input data. 

The idea of the CNN2D images method is to represent the complete distribution of the data, given the specific uncertainties.
We then organize the data, including the errors, in terms of a matrix whose columns correspond to the features ($i$), and the rows correspond to the values of distribution function of the input data.
To be specific, we discretize the range of input values in terms of bins $x_i \to x_i^\rho$, with $\rho=1,2,\ldots, \texttt{n\_rows}$, and then define values of the pixels of the CNN2D images as:
\begin{equation}
    \label{Eq:DefImage}
    P_{\rho, i} = \frac{1}{\sqrt{2 \pi \sigma_i^2}} \, 
    \exp \left[ -\frac12 \frac{(x_i -x_i^\rho)^2}{\sigma_i^2}  \right] \; .
\end{equation}
The left panel of Fig. \ref{fig:input data} illustrates the input data for this model.

The input shape of this model is $(\texttt{n\_rows}, n)$, where the number of rows \texttt{n\_rows} is one among other hyperparameters of the images that must be chosen. More details about the construction of the images can be found in  Appendix \ref{app:images tests}.
The standard network for the baseline training set (Table \ref{table:baseline sf}) consists of three convolution kernels with shapes $(5, 5)$, $(3, 3)$ and $(3, 3)$ with $32, 64$, and $64$ filters, respectively. The convolution layers are followed by \texttt{MaxPooling} layers with kernels with size and stride of $(2, 2)$, except for the first layer where the shape and stride of the kernel were chosen to be such that the output shape of this layer is always (10, 10), i.e, it depends on the shape of the input image.
%This configuration is, therefore, a generalization of the CNN1D no-$\sigma$ case for two dimensions.

The idea of representing a data vector with errors in terms of an image can be generalized for scientific data that is given in terms of pairs $\{ x_i \pm \sigma_{x_i} , y_i \pm \sigma_{y_i} \}$. 
In that case, the multivariate probability distribution associated with the uncertainties $\sigma_{x_i}$ and $\sigma_{y_i}$ mean that each data point $i$ is ``spread out'' both in the horizontal (rows, $x$) as well as the vertical (columns, $y$) directions. 
A CNN where a 2D data set (including uncertainties) is represented by images was used recently to classify supernovas  \citep{2021arXiv210604370Q}.
 % acho que cabe incluir um pseudo código.

\section{ML confronts maximum likelihood}
\label{results gauss}

We now present the results for the classification of curves in the two classes (smiley or frowny), in the presence of Gaussian noise in the input data. As discussed above, we generate the parabolic curves by sampling random parameters ($a, \ b$ and $c$) for the two classes, and then computing the $n$ features of the curves ($\bar{x}_i$), according to Eq. (\ref{Eq:xs}-\ref{Eq:xf}). 
The next step is to add noise to those features, according to Eqs. (\ref{Eq:xdx}-\ref{Eq:dist}).
It is important to stress that, just like in real experiments, the values $g_i$ are stored, which allows us to keep track of the error bar for each data point, $g_i \sigma_0$. 
However, for each object neither the underlying ``true" curve, $\bar{x}_i$, nor the random noise, $\delta x_i$, are known a priori: we only have access to the measurement, $x_i = \bar{x}_i + \delta x_i$, and the noise levels $g_i \sigma_0$ (the ``error bars'').

%Unless noted otherwise, our networks are trained with a sample of $m = 2 \times 10^5$ objects, a number of features $n = 20$, nominal error $\sigma_0 = 0.5$, and noise dispersion parameters $\bar{g} = 0.6$ and $\Delta g  = 0.5$. 
In order to study how the performances of the different networks change as a function of the parameters of the baseline dataset, we will let some individual parameters change, while the others are fixed to the values from Table \ref{table:baseline sf}.

\comment{
\begin{figure}[!htbp]
    \centering
    \includegraphics[width=1.0\linewidth]{smiley_frowny/parabolas_1.png}
    \caption{Parabolic curves generated according to the Smiley-Frowny model and the corresponding noisy measurements.}
    \label{fig:smiley frowny data and curves}
\end{figure}
}

\subsection{Varying the signal-to-noise ratio}

\begin{figure}[!htbp]
    \centering
    \includegraphics[width=0.8\linewidth]{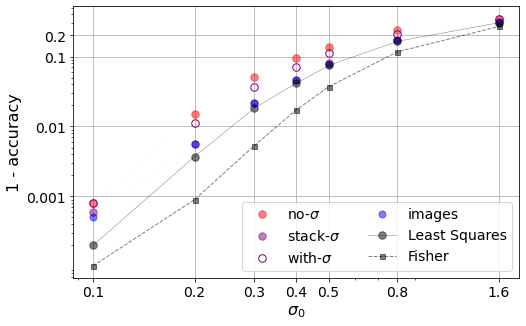}
    \caption{1 - accuracy as a function of $\sigma_0$ for fixed $\bar{g} =0.6, \Delta g = 0.5$ and $n = 20$.}
    \label{fig:vary sigma0 gauss}
\end{figure}

We start by investigating how the performance of the CNN classifiers, as well as the maximum-likelihood (least squares) method, depend on the nominal noise $\sigma_0$.
Fig. \ref{fig:vary sigma0 gauss} shows the accuracy of the classifiers. 
For very high nominal noise all the classifiers eventually fail, with accuracies of 50\%, and 
conversely, for very small noise all models are able to achieve a near-perfect accuracy. 
Therefore, as one should expect, in both limits ($\sigma_0 \to 0$ and $\sigma_0 \to \infty$) all methods become equivalent, since there is no information in the noise levels.

However, for intermediate levels of noise, the classification methods that make use of the information about the specific noise levels in the input data are able to achieve better performance than the CNN1D no-$\sigma$ method, which does not.
Moreover, for $\sigma_0 \gtrsim 0.3$, the CNNs that take into account the noise levels of the input data are able to reach accuracies which approach that of the maximum-likelihood classification. 
This means that when we pass the information about which features are more noisy and should be down-weighted, and which features have less noise and should be up-weighted, we allow the algorithms to learn about how to use those weights for their classifications.

From the perspective of training the CNNs, depending on the value of $\sigma_0$ we have added more regularization or reduced the number of layers. 
Since noisier objects tend to overfit the network, for large values of $\sigma_0$ the standard architecture for the baseline data set may not be optimal, and sometimes the models do not converge properly.

In particular, we note that the CNN1D stack-$\sigma$ method has a performance that is very similar to the CNN2D images method. 
This is perhaps to be expected, since the noise is Gaussian, and in the absence of skewness or kurtosis the key information about the distribution is already encoded in the standard deviation, $g_i \sigma_0$. 
For more general PDFs, where it is not possible to summarize the shape of the distribution in terms of a single parameter, using the CNN2D images approach might be an interesting alternative.
Alternatively, one could think of generalizing the CNN1D with-sigma models to multiple channels, each one containing the different momenta of the underlying distribution functions.

Still looking at Fig. \ref{fig:vary sigma0 gauss}, we see that all three models that include the errors have better accuracies than CNN1D no-$\sigma$, but CNN1D with-$\sigma$ is slightly worse than the other two. This difference can be explained by a fundamental difference in the form errors are provided to the models. While the point-wise signal-noise association is preserved in the input of CNN1D stack-$\sigma$ and CNN2D images, the same does not happen for CNN1D with-$\sigma$, causing the natural association to be lost. The model may eventually recover such association, but dismissing the association only adds unnecessary difficulties to the model. Therefore, in what follows we opted to discard the CNN1D with-$\sigma$.

\begin{figure}[!htbp]
    \centering
    \includegraphics[width=0.8\linewidth]{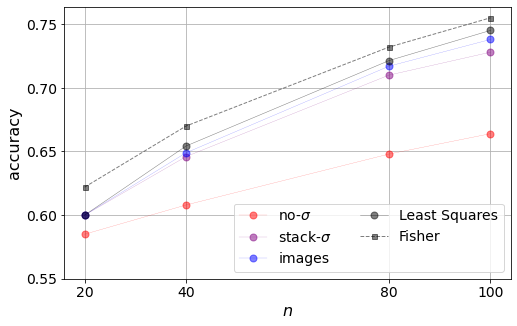}
    \caption{Accuracy as a function of the number of features $n$ for fixed $\bar{g} = 0.6, \Delta g = 0.5$ and $\sigma_0 = 3.2$.}
    \label{fig:vary n gauss}
\end{figure}

The typical values of $\sigma_0$ for which we see a significant difference between the models that include or do not include the uncertainties depend on the number of features $n$. 
If there is an abundance of points that characterize the curve, the classification become easier, and $\sigma_0$ should be larger in order to cause some confusion between the classes. In other words, it is equivalent to either lower the overall level of noise ($\sigma_0$) or to increase the number of features ($n$).
This is shown in Fig. \ref{fig:vary n gauss}, where we evaluate the performance of the models as we increase the number $n$ of features, for a fixed $\sigma_0 = 3.2$.
As we grow the number of features, all methods become more efficient, but the CNN1D method without errors clearly underperforms the other methods.

As the number of features grows, we are, once again, able to use less regularized architectures because, as we increase the overall signal, noise becomes less important and the networks become less sensitive to overfitting. 
Again, this is in agreement with the case where we vary $\sigma_0$.
%When we vary the number of features, the shape of the input matrices in CNN2D images changes as well.
%However, the shape and stride of the kernels in the first \texttt{MaxPooling} layer for the different $n$ were chosen to be such that the output shape of this layer is always (10, 10). For example, if $n=40$, the input shape is $(40, 20)$, then the \texttt{MaxPooling} kernel shape and stride is $(4, 2)$.

\subsection{Varying $\Delta g$}

\begin{figure}[!htbp]
    \centering
    \includegraphics[width=0.8\linewidth]{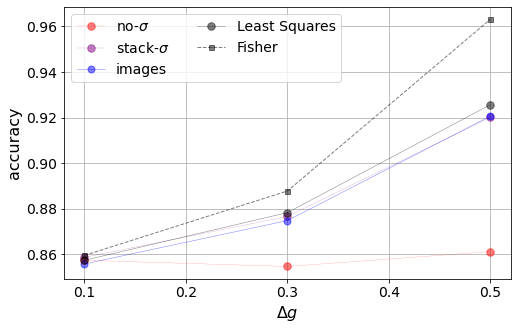}
    \caption{Accuracies for $\Delta g = 0.1, 0.3, 0.5$ with fixed $\bar{g} = 0.6$, $n=20$ and $\sigma_0 = 0.5$.}
    \label{fig:vary g gauss}
\end{figure}

Figure \ref{fig:vary g gauss} shows the performance of the classifiers as the noise variance parameter $\Delta g$ changes. 
This plot shows that, as we increase the relative difference between the noise levels of the data points, that information becomes more critical to the classification. 
That information is naturally used in the least squares method, and is also learned by the CNNs that are provided the error bars, but the CNN that is only provided the input data is unable to harness that information to improve the classification.

\subsection{Probability Output}

We now address the question about the quality of the ML classifiers as compared with an MCMC approach. 
For each object we explore the likelihood function, Eq. (\ref{Eq:lik}). 
We use flat priors for the parameters, which were free to vary inside the ranges $a \in [-3,3]$, $b \in [-5,5]$ and $c \in [-5,25]$ -- i.e., more than 8 standard deviations for the parameter $a$, and about three standard deviations away from the means for the parameters $b$ and $c$.

The goal of the MCMC is to compute the posterior probability that the maximum likelihood classification for each individual object is correct. 
In order to compute that probability, we count the fraction of points in the chains that are assigned the correct class of each object.
Since each object is also assigned a class by the three different ML methods, together with their respective confidences, we can plot the ML confidence as a function of the MCMC probability for all objects in the test sample. 

\begin{figure*}[!htpb]
    \centering
        \includegraphics[width=1.0\linewidth]{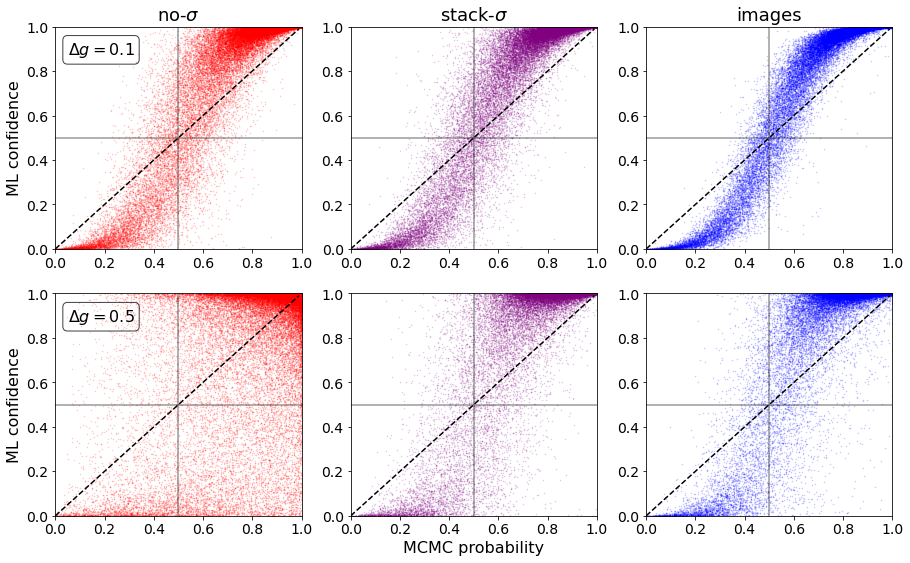}
    \caption{CNNs confidence \textit{vs} MCMC probability for the classification of 50 000 smiley objects. 
    The noise models used for this test had $\sigma_0 = 0.5$, $n=20$, with noise dispersion given by $g \in [0.5, 0.7]$ (nearly homoscedastic case, top), or by $g \in [0.1, 1.1]$ (heteroscedastic case, bottom).}
    \label{fig:MCMC vs ML}
\end{figure*}

\begin{figure*}
        \centering
        \begin{subfigure}%{0.5\textwidth}
            \centering
            \includegraphics[width=.495\linewidth]{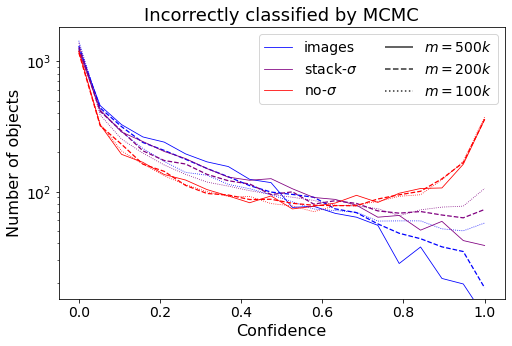}
            %\label{fig:su  %\caption{A subfigure}
        \end{subfigure}%
        \begin{subfigure}%{0.5\textwidth}
            \centering
            \includegraphics[width=.495\linewidth]{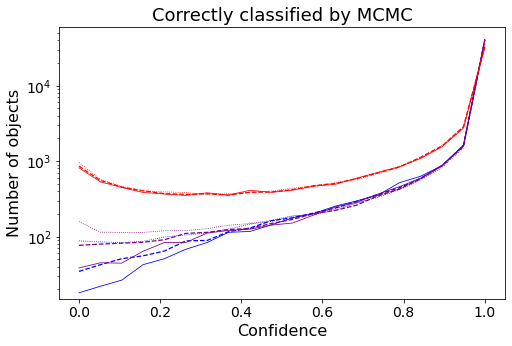}
            %\caption{A subfigure}
            %\label{fig:sub2}
        \end{subfigure}
    \caption{Confidence of the CNNs in the classification of the smiley objects which were incorrectly (left) and correctly (right) classified by MCMC, in the case where $g\in [0.1,1.1]$. 
    The model parameters are the same as the ones used for Fig. \ref{fig:MCMC vs ML}, but here we also vary the size of the training set.
    Solid lines correspond to training sets with $m = 500 \times 10^3$ instances, dashed lines $m = 2 \times 10^5$ and dotted lines $m = \times 10^5$. The curves are the mean values over multiple training realizations.}
    \label{fig:confidence}
    \end{figure*}

This comparison is shown in Figures \ref{fig:MCMC vs ML} and \ref{fig:confidence}. 
In Fig. \ref{fig:MCMC vs ML} we show the CNN confidence and the MCMC probability for $5\times 10^4$ objects in the smiley (positive) class, using the baseline dataset. 
Objects with MCMC probability $>0.5$ are classified in the correct class, and when the probability is $\leq 0.5$ the maximum likelihood classification fails. 
The same applies for the CNNs: a  confidence of 0.5 marks the threshold between correct and wrong classification. 

The most revealing aspect of Fig. \ref{fig:MCMC vs ML} is that, when the noise levels of all input data points are nearly the same (nearly homoscedastic case, $\Delta g = 0.1$, top row), the three ML methods hold a tight correlation between the confidence of the classification and the MCMC probability that the classification is correct.
However, when the data points have significantly different levels of noise (heteroscedastic case, $\Delta g = 0.5$, bottom row), if those uncertainties are not passed on to the CNN (as is the case of the CNN1D no-$\sigma$ model, left panel), then there is basically no correlation between the ML confidence and the MCMC probability. But when the input data uncertainties are part of the information provided to the CNNs, a clear correlation appears between the ML confidence and the probability. In fact, the ML confidence is well fitted by the sigmoid function:
\begin{eqnarray}
    c_\text{ML} (p) \sim \frac{p^2}{p^2 + (1-p)^2} \; .
\end{eqnarray}
This function can be easily inverted, which allows us to define:
\begin{eqnarray}
    p_\text{ML} = \frac{c_\text{ML} - \sqrt{c_\text{ML}-c_\text{ML}^2} }{2 c_\text{ML} - 1} \; .
\end{eqnarray}
A measure of the goodness of this fit can be computed with the mean squared error (MSE) metric:
\begin{equation}
    \text{MSE} = \frac{1}{m_\text{test}} \sum_{j=1}^{m_\text{test}} \left[ p_\text{ML}(j) - p(j) \right]^2 \; ,
\end{equation}
where $m_\text{test}$ is the number of objects in the test sample, and $p(j)$ is the probability (according to the MCMC) that the classification of object $j$ is correct.

We obtain that, for $\Delta g = 0.1$, the MSE for all methods is below 0.01, with the CNN2D images method performing slightly better at 0.0026, compared with 0.0059 for CNN1D no-$\sigma$ and 0.0052 for the CNN1D stack-$\sigma$ model (we use a sample of $m_\text{test} = 5\times 10^5$ objects in order to compute this statistic).
When we increase the noise dispersion to $\Delta g = 0.5$, the CNN1D no-$\sigma$ fails to fit the sigmoid, with an MSE of 0.0386, whereas the CNN1D stack-$\sigma$ and CNN2D images methods still performing well, below 0.01 -- see Table \ref{table:mse}.

\begin{table}[!htbp]
\caption{MSE of the cases shown in Fig. \ref{fig:MCMC vs ML}.}
\label{table:mse}
\centering
\begin{tabular}{c c c c c c}
\hline
\hline
 \ & CNN1D no-$\sigma$ & CNN1D stack-$\sigma$ & CNN2D images \\
\hline
 $\Delta g = 0.1$ & 0.0059 & 0.0052 & 0.0026 \\
 $\Delta g = 0.5$ & 0.0386 & 0.0087 & 0.0064\\
\hline
\end{tabular}
\end{table}

The results summarized in Figures \ref{fig:MCMC vs ML} and \ref{fig:confidence}, together with Table \ref{table:mse}, mean that, when the noise levels of all the input data are the same (or nearly the same), then the ML methods are able to estimate the quality of the classification (the confidence) in a way that works as a proxy for the probability that this classification is correct.
On the other hand, if different data points have varying levels of noise, then it becomes essential to pass that information on to the networks. When that information is hidden from the network, as happens for CNN1D no-$\sigma$, then the method loses its ability to provide a confidence that is significantly correlated with the probability for that classification -- see the bottom left panel of Fig. \ref{fig:MCMC vs ML}.
However, if we pass the noise properties of the data as information to the CNNs, then we allow those networks to reconstruct confidences that are tightly correlated with the probabilities for the classification.

Another way to visualize the predictive power provided by the noise information is to take the points shown in Fig. \ref{fig:MCMC vs ML}, and separate them into objects that are correctly classified by the MCMC (probability $> 0.5$), and those that are incorrectly classified. 
In Fig. \ref{fig:confidence} we show the resulting distribution of objects as a function of the ML confidence for the incorrectly classified (left panel) and correctly classified (right panel) objects.
We also show how that confidence varies with the size of the training sets ($m = 1, \ 2$ and $5 \times 10^5$ objects). 
It is immediately clear that the CNN that is blind to the information about uncertainties is unable to pick out the more noisy objects, and as a result it assigns high confidences to objects in the wrong class much more often than the other methods. 
Furthermore, for the objects that are correctly classified by the MCMC (right panel), the CNN1D no-$\sigma$ method tends to assign lower probabilities to more objects, which again is a result of those methods being unable to weigh features by their uncertainties (notice that the log-scale plot makes it a bit difficult to see that the number of correctly classified objects with the highest confidence is significantly lower for the CNN1D no-$\sigma$ method).

\subsection{Varying the training set size}

An important issue that appears as we increase the dimensionality of the system by including additional parameters related to noise is the size of the training set that is needed for the network to converge.
We have evaluated the performance of all classifiers as a function of the size of the training set for our baseline model, with $n = 20$, $\sigma_0 = 0.5$, $\bar g = 0.6$ and $\Delta g = 0.5$. 
We froze the same architecture that was used with $m = 2 \times 10^5$ objects in the training set, and re-trained the network with different sizes in order to see how its performance degraded or improved as we lower or increase the number of objects. We also analyse how much sensitive the model becomes to the initial seed as the training set size decreases. For lower number of objects, it is harder for the model to converge. This is not the case for larger training sets, where the model converges to very similar results with different initial seeds.
%For this study, we generated a data set with $m = 10^6$ objects, half smiley, half frowny. We then divided this data set into slices, each one corresponding to a training set.
Of course, it is still possible to improve the accuracy for larger/lower $m$ if one reduces/increases the complexity of the network.

    \begin{figure*}
        \centering
        \begin{subfigure}%{0.5\textwidth}
            \centering
            \includegraphics[width=.495\linewidth]{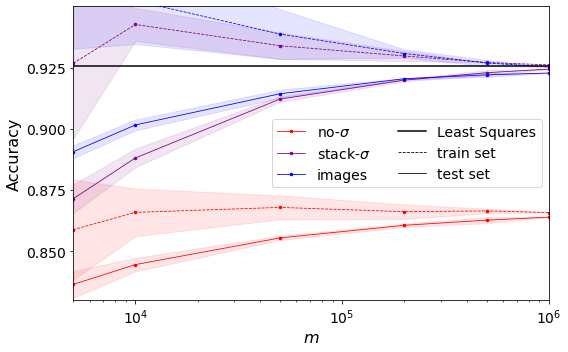}
            %\label{fig:su  %\caption{A subfigure}
        \end{subfigure}%
        \begin{subfigure}%{0.5\textwidth}
            \centering
            \includegraphics[width=.495\linewidth]{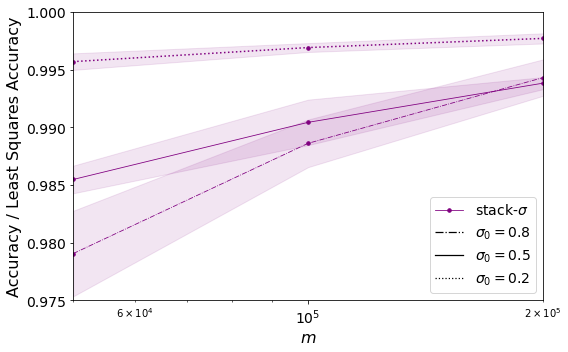}
            %\caption{A subfigure}
            %\label{fig:sub2}
        \end{subfigure}
    \caption{\textit{Left}: Accuracies as a function of the training set size $m$. Solid and dashed lines correspond to test and train set, respectively. The curves are the median value of multiple realizations and the shaded region covers the interval median +/- $\sigma_\text{NMAD}$ \eqref{eq:sigma nmad}; \textit{Right}: Ratio between the accuracy of CNN1D stack$-\sigma$ and the accuracy of least squares as a function of the number of training set instances for $\sigma_0 = 0.2, 0.5, 0.8$.}
    \label{fig:vary m}
    \end{figure*}

The left panel of Fig. \ref{fig:vary m} shows the accuracy for both train (dashed lines) and test (solid lines) sets of the models, as we increase the number of training instances. 
This means that, as we take $m\to \infty$, we have minimized epistemic error for each ML method, and all that remains is the impact of aleatoric errors on the different models.
The lines correspond to the median value of multiple realizations, and the error bars (shaded regions) correspond to ``normalized median absolute deviation'' $ \sigma_\text{NMAD}$, which is defined by:
\begin{equation}
\label{eq:sigma nmad}
    \sigma_\text{NMAD} =  1.4826 \times \text{median}(| y_i - \text{median}(y) |) \, .
\end{equation}
This measure of the width of a PDF reduces to the variance in the case of a mono-variate Gaussian distribution, but is less affected by the tails of the distribution.

We see that, for the training set sizes that we analyzed, no amount of training data is sufficient for CNN1D no-$\sigma$ to come close to the performance of the models that include the information about uncertainties\footnote{For extremely small training sets it may be possible that the lower complexity of the methods without noise information makes them superior to the ones that include noise.}.
Notice that our toy model allows us to generate an arbitrarily large number of objects, while data augmentation usually employs some fixed set of objects and then adds an artificial amount of noise to the input data of those objects. 
Therefore, the larger training sets of Fig. \ref{fig:vary m} are composed of objects which behave exactly like the ones in the validation and test sets, while data-augmented training sets are made up of a mix of original objects as well as objects which behave in a fundamentally different way compared with the original ones. 
As a result, the performance of a model trained on a set of $m$ original objects is always superior to that of a model trained on a set of $m$ objects that was created with the help of data augmentation techniques. 
Hence, this result also proves that data augmentation techniques cannot possibly overcome the deficit of neglecting the information about uncertainties. 

Both CNN2D images and CNN1D stack-$\sigma$, even if the former presents higher complexity than the latter, have similar performances, specially when considering the intrinsic model variations (shaded regions in Fig. \ref{fig:vary m}). 
Moreover, the amount of data required for the models to converge to some accuracy depends on how noisy the data is:  when the data is more noisy, the models are more likely to display overfitting. 
This is shown in the right panel of Fig. \ref{fig:vary m}, where we plot the accuracy of the models as a function of size of the training, for different values of $\sigma_0$ (we only show the results for the CNN1D stack-$\sigma$, for clarity, but all methods behave in essentially the same way.) 
Notice that in this plot we normalized the accuracy of the CNN classification by the accuracy of the least squares (maximum likelihood) classification, in order to highlight the fact that the difference between those accuracies is more pronounced for larger values of the nominal uncertainty $\sigma_0$.

\section{Poisson Noise}

The Poisson distribution is particularly interesting because the uncertainty of the measurement can be deduced from the measurement itself. 
In that sense, it would be redundant to provide the noise information: if a feature has a measured value $x_i$, a good estimator for the noise is already given by $\sqrt{x_i}$. 
Hence, we expect that the information of the uncertainty is already encoded in the value of the measurement and, therefore, it does not make any difference to use as input only the measurement, with CNN1D no-$\sigma$, or to use the measurement and its associated error bar, like we do with CNN1D stack-$\sigma$ or CNN2D images.

\comment{
\begin{figure*}
        \centering
        \begin{subfigure}%{0.5\textwidth}
            \centering
            \includegraphics[width=.45\linewidth]{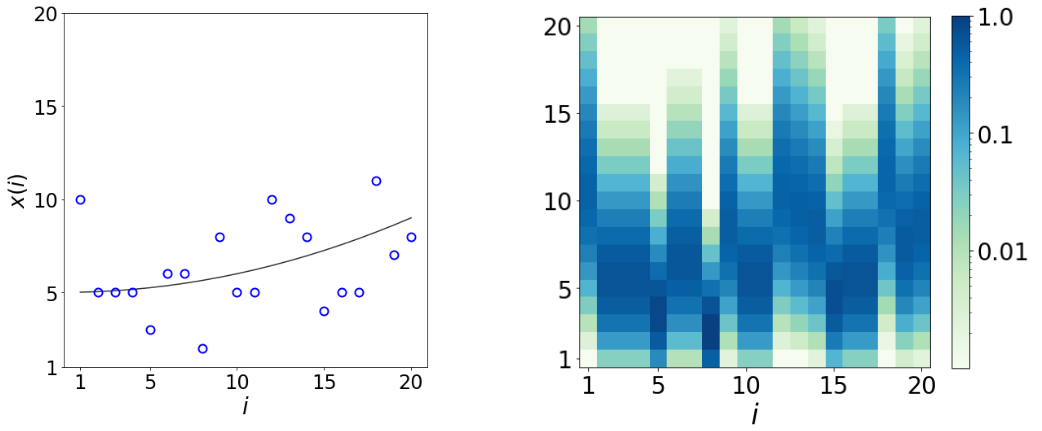}
            %\label{fig:su  %\caption{A subfigure}
        \end{subfigure}%
        \begin{subfigure}%{0.5\textwidth}
            \centering
            \includegraphics[width=.5\linewidth]{poisson_image.png}
            %\caption{A subfigure}
            %\label{fig:sub2}
        \end{subfigure}
    \caption{Smiley-frowny curves generated with Poisson noise.}
    \label{fig:vary m}
    \end{figure*}
}

\begin{figure}[!htbp]
    \centering
    \includegraphics[width=1.0\linewidth]{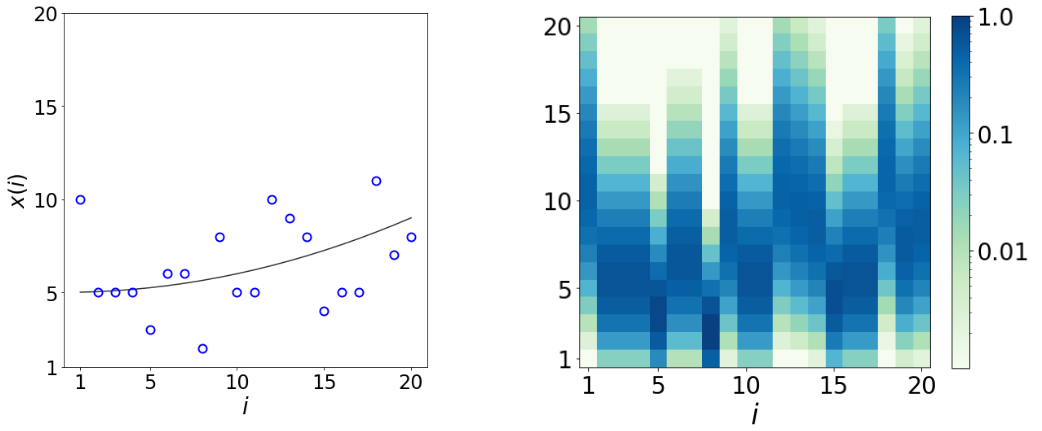}
    \caption{Smiley-frowny curves generated with Poisson noise.}
    \label{fig:sf poisson}
\end{figure}

In order to work with Poisson noise, it is convenient to employ measurements that have values closer to one, in order to reinforce the skewness of the distribution.
Therefore, we used an adapted version of the smiley-frowny model to study the Poisson case. 
We lowered the mean value of the parameter $c$ of the parabolic curves to get mean values closer to one, and also reduced its standard deviation to avoid measurements with negative values. 
The few objects which happened to display any feature with negative values were discarded. 
Finally, since the noise in this case turns out to be relatively small, we also chose different values of $c$ for smiley and frowny objects, so we could shuffle the curves more efficiently and prevent the machine from distinguishing between both classes by their absolute values instead of the curvature. Table \ref{table:parameters2} presents the distribution of parameters used to test the Poisson noise model, and
Fig. \ref{fig:sf poisson} shows the input data for CNN2D images with $n = 20$, which represent the measurements.

\begin{table}[!htbp]
\caption{Parameters of the normal distributions from which the coefficients of the parabolic curves are sampled - Poisson version.}     
\label{table:parameters2}
\centering
\begin{tabular}{l c c c c c c}
\hline
\hline
Coefficient & $a_\smile$  & $a_\frown$ & $b$ & $c_\smile$ & $c_\frown$ \\    
\hline
   $\mu$  & $1$ & -1 & 0 & 3 & 4 \\
   $\sigma$  & 0.25 & 0.25 & 1 & 0.5 & 0.5 \\
\hline
\end{tabular}
\end{table}

\begin{figure}
    \centering
    \includegraphics[width=0.8\linewidth]{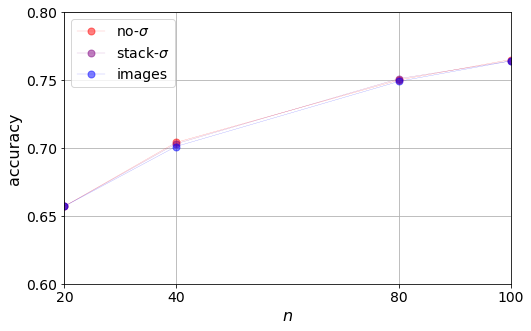}
    \caption{Accuracy as a function of the number of features $n$ in the Poisson version of the smiley-frowny model.}
    \label{fig:vary n poisson}
\end{figure}

In the case of Poisson, we do not have a parameter such as $\sigma_0$, as we had in the Gaussian noise case. However, we can vary the level of noise by increasing or decreasing the number of features, $n$. 
Fig. \ref{fig:vary n poisson} shows the dependence of the performance with $n$ for the Poisson case -- compare it with Fig. \ref{fig:vary n gauss}, for the Gaussian noise model.

These results show that the accuracy improves as $n$ increases, but all models have basically the same accuracy. What it means is that including or not the uncertainty is irrelevant when the noise is drawn from a Poisson distribution: indeed, in that case there is no additional information in the noise levels (error bars) that is not already present in the measurements (the input data).

\section{The waveform dataset}
\begin{figure}
    \centering
    \includegraphics[width=0.75\linewidth]{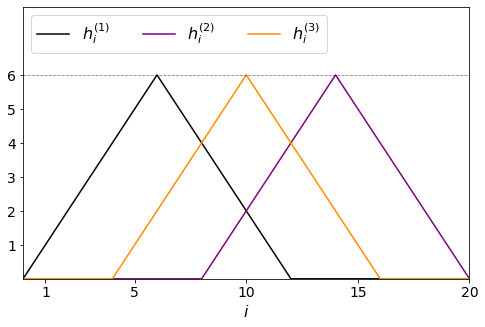}
    \caption{$h$ functions of the waveform dataset.}
    \label{fig:waveforms h}
\end{figure}

We now apply our methods to the publicly available waveform data set\footnote{Competition scores: https://www.openml.org/t/58} \citep{reason:BreFriOlsSto84a}.
%(Breiman,L., Friedman,J.H., Olshen,R.A., and Stone,C.J. (1984). Classification and Regression Trees. Wadsworth International, pp 49-55, 169.). 
That data set consists of three different combinations of three functions:
\begin{align}
    &\text{class 1}: x^{1}_i = u \, h^{(1)}_i + (1 - u) \, h^{(2)}_i + \delta x^{1}_i \nonumber \\
    &\text{class 2}: x^{2}_i = u \, h^{(1)}_i + (1 - u) \, h^{(3)}_i + \delta x^{2}_i \nonumber \\
    &\text{class 3}: x^{3}_i = u \, h^{(2)}_i + (1 - u) \, h^{(3)}_i + \delta x^{3}_i \label{Eq:waveforms},
\end{align}
where $i = 0, 1, \dots, 20$ labels the $n = 21$ features, and the three ``parent'' waves are shown in Fig. \ref{fig:waveforms h} -- from left to right, $h^{(1)}$, $h^{(3)}$ and $h^{(2)}$. The random variable $u$ is drawn from a Uniform distribution, $u \in [0, 1]$.

There are two versions of the waveform dataset available in the UCI repository\footnote{Link to version 2: https://archive.ics.uci.edu/ml/datasets/waveform+database+generator+(version+2)}, which provides a data folder with a data generator code written in \texttt{C} and also a document with $5000$ waveform instances.
The versions differ from each other on the available information. The first version contains only the $21$ features and the labels. The second version contains, in addition to the $21$ features and labels, the $19$ noise values $\delta x_i$ for $i \in [1, 20]$. For $i = 0$ and $i = 21$, the value of the wave is zero for all three combinations (classes) (see Figure \ref{fig:waveforms h}), which means that $x_{0, 21} = \delta x_{0, 21}$. Therefore, explicitly adding $\delta x_i$ to the data set is redundant.

The task is to classify objects according to these three types, using the $n=21$ measurements and uncertainties.
In the original versions, the noises $\delta x_i$ from the waveform data set were generated from the same normal distribution $\mathcal{N}(\mu = 0, \sigma = 1)$. 
The data set actually included the noise itself, $\delta x_i$, instead of the standard deviation of the Gaussian distribution from which those $\delta x_i$ were sampled.
Since the noise is sampled from the same PDF for all points, there is no value in informing the networks about that noise.
For this reason, we adapted the code that generates those features in order to have noises with different variances.
Just as in the original waveform data set, we draw the uncertainties $\delta x_i$ from normal distributions:
\begin{equation}
    \mathcal{N}(\mu = 0,\ \sigma = g_i \cdot \sigma_0) ,
\end{equation}
but in our modified version the noise levels $g_i$ are sampled from a uniform distribution with $\bar{g} = 0.8$ and $\Delta g = 0.5$, i.e., $ g_i \in [ 0.3 , 1.3] $.

We have computed the performance of the three ML classifiers, CNN1D no-$\sigma$, CNN1D stack-$\sigma$, CNN2D images, as well as the classification using the maximum likelihood\footnote{The class was assigned by choosing the function (Eqs. \eqref{Eq:waveforms}) which fits the data with the lowest $\chi^2$.}, in the following situations: first, we kept $\Delta g=0.5$ fixed, and varied the nominal error parameter $\sigma_0$. Second, we fixed $\sigma_0 = 1.25$ and varied the noise dispersion parameter $\Delta g$.

\begin{figure*}
        \begin{subfigure}
            \centering
            \includegraphics[width=.48\linewidth]{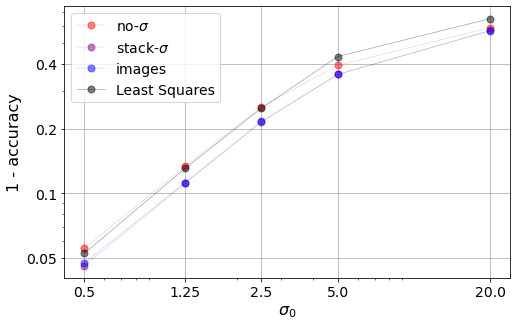}
            %\label{fig:su  %\caption{A subfigure}
        \end{subfigure}%
        \begin{subfigure}%{0.5\textwidth}
            \centering
            \includegraphics[width=.48\linewidth]{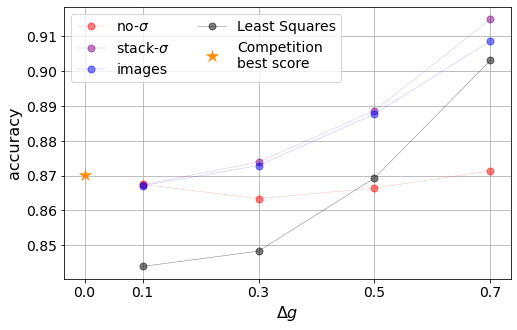}
            %\caption{A subfigure}
            %\label{fig:sub2}
        \end{subfigure}
    \caption{Results of the classifications applied to the modified waveform data set. \textit{Left}: fixed $\Delta g = 0.5$, varying $\sigma_0$; \textit{Right}: fixed $\sigma_0 = 1.25$, vary $\Delta g$. For both cases we take $\bar{g}=0.8$, thus $\bar{\sigma}_0 = 0.8 \, \sigma_0$. The standard waveform data set has $\bar\sigma_0 = 1$ and $\Delta g = 0$, which corresponds to extending the curves on the right panel to $\Delta g \to 0$. For comparison, the star symbol in the right panel shows the best accuracy ($0.8702$) obtained in the waveform classification competition (where $\Delta g = 0$).}
    \label{fig:wf vary sigma0 and vary g}
    \end{figure*}

The results are shown in Fig. \ref{fig:wf vary sigma0 and vary g}.
They resemble closely those shown in Fig. \ref{fig:vary sigma0 gauss} and in Fig. \ref{fig:vary g gauss}, and show that the CNNs that take into account the different levels of noise are superior to the method that discards that information, regardless of whether the data is overall less or more noisy (lower or higher values of $\sigma_0$).
The right panel of Fig. \ref{fig:wf vary sigma0 and vary g} shows that, as $\Delta g$ grows and the information about which data points are more or less noisy becomes increasingly relevant, the methods that can account for these differences in signal-to-noise outperform by far the CNN1D no-$\sigma$ method, which is blind to those distinctions.
Notice, in particular, that in the limit $\Delta g \to 0$ we recover the original waveform version, where the noise has a Gaussian distribution with 
$\mathcal{N}(\mu = 0, \bar\sigma = \bar{g} \cdot \sigma_0 = 1)$.

%We see from the confusion matrices in Figure \ref{fig:wf cm} that CNN2D images performs poorly on class 1 as compared to class 2 and class 3, which is not the case for Least Squares. {\color{purple}[Gastei um tempo para entender por que o CNN tem desempenho melhor nas classes 2 e 3, até que li o que a Natália escreveu com atenção. Eu não consigo ter a intuição -- talvez porque não tenho a intuição do least squares ... Em todo caso, se formos manter  essas matrizes, deveríamos ter alguma explicação para o que está na última frase? A figura da confusion matrix deveria ser inserida depois da figura com os gráficos (fig 14) -- no momento aparece ref. para fig 14 antes de para a fig. 13 ]}
\comment{
\begin{figure}[!htbp]
    \centering
    \includegraphics[width=0.95\linewidth]{waveform/WF_cm.png}
    \caption{Confusion matrix of the modified waveform dataset ($\sigma_0 = 1.25, \Delta g = 0.5$) for Least Squares (left) and CNN2D images (right).}
    \label{fig:wf cm}
\end{figure}
}

The main difference with respect to the smiley-frowny model is that now the maximum likelihood classification can be worse than the CNNs, since the latter are able to pick out the distinguishing global features of the parent waveforms.
This is because the maximum likelihood classification is derived from a point-wise fit, such that the input data points contribute to the fit independently from each other. The ML methods, on the other hand, are able to relate and combine different data points in order to detect the shape of the object, resulting in a more robust classification.

\comment{
\begin{table}[!htbp]
\caption{Accuracies of the waveform classification.}             % title of Table
\label{table:coeff parameters}      % is used to refer this table in the text
\centering                          % used for centering table
\begin{tabular}{c c c c}        % centered columns (4 columns)
\hline
\hline                 % inserts double horizontal lines
$\sigma_0$ & no-$\sigma$ & stack-$\sigma$ & images \\    % table heading 
\hline                        % inserts single horizontal line
   1.0 & 0.863 & 0.877 & 0.877 \\      % inserting body of the table
   2.0 & 0.745 & 0.769 & 0.768 \\
\hline                                   %inserts single line
\end{tabular}
\end{table}
}

\comment{
\begin{table}[!htbp]
\caption{$g$ values description.}             % title of Table
\label{table:parameters3}      % is used to refer this table in the text
\centering                          % used for centering table
\begin{tabular}{c c c c}        % centered columns (4 columns)
\hline
\hline                 % inserts double horizontal lines
$\Delta g$ & $g_\text{min}$ & $g_\text{max}$ \\    % table heading 
\hline                        % inserts single horizontal line
   0.1 & 0.7 & 0.9 \\      % inserting body of the table
   0.3 & 0.5 & 1.1 \\
   0.5 & 0.3 & 1.3 \\
   0.7 & 0.1 & 1.5 \\
\hline                                   %inserts single line
\end{tabular}
\end{table}
}

\section{Discussion and Conclusion}

In this paper we address the value of the information about noise in input data for Machine Learning (ML) methods. 
We have shown that, when a data set includes not only the (noisy) measurements of the features, but also the information about the underlying distribution functions that generated that noise, CNNs are able to learn about the context of that noise, improving the performance of classification tasks and reaching ``optimality'', defined here in terms of a maximum likelihood approach.

In order to prove this statement we created a toy model for two classes (the ``smiley'' and ``frowny'' parabolic curves), and a model for input data noise that realizes the typical process of measurement.
Each object was generated from parameters that obey a random process, allowing us to build arbitrarily large sets that we can use to train, validate and test our methods.
Noise, on the other hand, was also generated by means of a random process, but in such a way that each data point (feature) has a noise that is drawn from a different PDF, whose dispersion is known: this is the ``error bar'' associated with each feature. 
This is exactly what takes place in a laboratory: the experimenter not only takes the measurement, but also assesses the uncertainties of each measurement -- which are typically not all identical.

As a result, not only the objects in our two classes have known underlying distributions, but each object can be classified using a maximum likelihood approach. 
This creates a standard against which the ML methods can be compared, as well as the concept of an ``optimal'' accuracy for the classifiers. 
Notice that optimality, defined in this sense, has to be used with great care: ML methods can outperform maximum likelihood estimators when there are non-local patterns in the data that can serve to distinguish the objects. 
However, in our toy model we precluded any such patterns from appearing, since we limited the distinction between the two classes to a single parameter -- the curvature. 
It is in that sense that we can define optimality.

Our main result is that, when the information about data noise is passed on to a CNN, it can learn how to use the different levels of noise to weigh the input data. 
This leads to improvements in the performance of the classifiers, in such a way that the accuracy of the ML classification approaches that of the optimal (maximum likelihood) estimator.
In fact, the more the noise levels vary from point to point (as controlled by the noise dispersion parameter $\Delta g$), the better the performance of the CNNs that included the noise level information compared with the CNN that did not.

Moreover, we showed that, when the levels of input data noise are not all identical, the confidence of the ML method that is ignorant about those specific noise levels becomes uncorrelated with the underlying cumulative distribution function -- see Fig. \ref{fig:MCMC vs ML}.
However, when the noise levels are provided as additional data inputs to the CNNs, the resulting confidence of the classifiers can again be mapped onto the MCMC probability for the classification. 
Although that mapping is noisy, Fig. \ref{fig:MCMC vs ML} shows that the CNNs seem to be using the information about the different levels of noise in the input data to reconstruct what is, in effect, a proxy for the likelihood function.

We have further tested CNNs with and without the noise level information using a slightly modified version of the waveform data set \citep{reason:BreFriOlsSto84a}. 
Just as happened for the smiley-frowny model, including the information about the different noise levels improves the accuracy of the classification, by an amount that becomes larger as we increase the noise dispersion parameter $\Delta g$.
We also computed the classification of objects in that data set using a maximum likelihood approach, however in that case the global patterns of the objects (in this case, the two peaks at known positions) can be detected by the CNNs, hence in some instances the maximum likelihood method was inferior to the CNNs.
Nevertheless, when the information about noise levels is included in the CNNs, they always overperform the maximum likelihood classification -- see Fig. \ref{fig:wf vary sigma0 and vary g}.

We also checked that the noise levels are only relevant when they provide information that is not already included in the data set itself. 
In order to show this, we created a modified version of the smiley-frowny model whose features are numbers drawn from a Poisson distribution. 
In that case, for each feature $x_i$ the noise levels are well approximated by $\sqrt{x_i}$, and therefore there is very little additional information being provided by adding those errors to the data set.
And indeed, what we find is that in this case the CNNs with and without error information have basically identical accuracies (after accounting, of course, for the different levels of complexity of the models.)

It is important to stress that ignoring the information about the different levels of noise in input data degrades the quality of ML classifiers in a way that cannot be offset by adding objects to the training set -- through, e.g., the use of data augmentation techniques. 
As can be seen in Fig. \ref{fig:vary m}, increasing the size of the training set, even in an ideal setup such as the one provided by our toy model, is not sufficient to allow the CNN without error information to achieve the accuracy level of the CNNs that include that information. 
In other words, the noise information is essential, and cannot be substituted or compensated. 
Moreover, as discussed in Section \ref{toy_model}, regularization techniques are also insufficient to compensate for the lack of information about the different levels of input data noise.

Finally, we ought to remark that our  conclusions were drawn in the context of a model inspired by scientific data, where tracking signal and noise are commonplace. However, in all areas of data science the issue of measurement is a key one: some data sets are more robust than others, and some data points are more reliable than others. Furthermore, estimations about the levels of noise in input data are often available to the data analyst not only in the hard sciences, but, e.g., in economic data or in the social sciences as well. 
What we have shown here is that providing these noise levels to ML methods adds significant information to the algorithms, improving the performance of classification or regression tasks in a way that cannot be compensated by  techniques such as data augmentation or regularization.

Python codes and documentation can be found at \url{https://github.com/nvillanova/smiley_frowny} .

% Acknowledgements should go at the end, before appendices and references

\acks{Many thanks to Eloi Pattaro, who first got two of the authors (N.R. and R.A.) started on this path.
We also thank Laerte Sodré Jr. for insightful conversations about Machine Learning applications in Astronomy. We thank Google Collaboratory for the GPU resources.
Financial support for this project was provided by CAPES, CNPq, and FAPESP through grants 2019/26492-3 (R.A.), 2017/25835-9 and 2015/22308-2 (N.H.).}


\newpage

\appendix
\section{Describing the images method}\label{app:images tests}

% Note: in this sample, the section number is hard-coded in. Following
% proper LaTeX conventions, it should properly be coded as a reference:

%In this appendix we prove the following theorem from
%Section~\ref{sec:textree-generalization}:

In this appendix we discuss the construction of the images in more details. The shape of the images are defined by a set of hyperparemeters, i.e., parameters that must be chosen before training the model. As mentioned in Section \ref{images section}, we discretize the features $x_i \to x_i^\rho$ and define the pixels of the images according to Eq. \eqref{Eq:DefImage}.

To set the height and width of the matrix, we define the variables:
\begin{align*}
    &\texttt{n\_rows}:\ \text{number of rows}\\
    &\texttt{n\_cols}:\ \text{number of columns}
\end{align*}
For simplicity, we set \texttt{n\_cols} = $n$, i.e., the number of features.

To define the boundaries of the image of each object $j\ (j = 1, \dots, m)$ in the data set, we use the variables \texttt{up\_bound} and \texttt{low\_bound}\footnote{For the tests with the Poisson version of the smiley-frowny model, we fixed the boundaries to range between $0$ and $20$ for all the objects.}:
\begin{align*}
    &\text{upper boundary} = \sum_i^n \frac{x_i^{(j)}}{n} + \texttt{up\_bound}\\
    &\text{lower boundary} = \sum_i^n \frac{x_i^{(j)}}{n} - \texttt{low\_bound} \; .
\end{align*}

\begin{figure}[!htbp]
    \centering
    \includegraphics[width=1.0\linewidth]{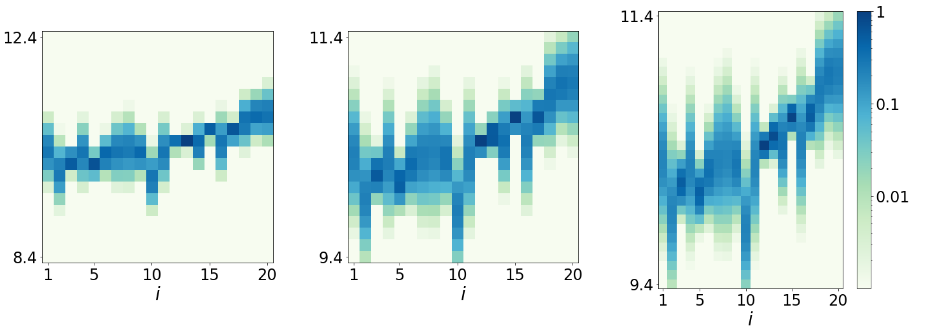}
    \caption{Input data of CNN2D images (smiley-frowny, $n=20, \sigma_0 = 0.2$) with different images hyperparamters $\{ \texttt{n\_rows}, \texttt{up\_bound}, \texttt{low\_bound} \}$. From left to right: $\{20, 2, 2\}, \{20, 1, 1\}, \{30, 1, 1\}$ (\texttt{n\_cols} = $n$ in all cases).}
    \label{fig:vary resolution}
\end{figure}

The choice of these hyperparameters might have a relevant impact on the performance of the model in some cases because these parameters define how much information of the data is being communicated through this representation.
%They must be chosen in such a way that the image is as dense as possible, i.e., the largest amount of information is represented in the most compact way.
To illustrate this, we show in Figure \ref{fig:vary resolution} a smiley curve with $n = 20$ and $\sigma_0 = 0.2$ represented in three images built with different hyperparameters. We see that these parameters control the ``resolution" of the curve, i.e., the number of pixels to bin $x_i$. More specifically, the width of the bins expressing the intervals for the values $x^\rho_i$ are given by:
%We see that if we set too small values for \texttt{up\_bound} and \texttt{low\_bound}, one may cut out a part of the curve. On the other hand, if \texttt{up\_bound} and \texttt{low\_bound} are too large, one may change \texttt{n\_rows} to control the ``resolution" of the curve, i.e., choose the number of pixels to bin $x_i$.
\begin{equation}
\label{eq:bin size}
    \Delta x = \frac{\texttt{up\_bound} + \texttt{low\_bound}}{\texttt{n\_rows}} \, .
\end{equation}

\section{Evaluating the images method}\label{app:vary threshold}

Here we perform a test with CNN2D images model to try to get some intuition on how the model is using the information of the uncertainties. The idea is to evaluate how the accuracy changes as we gradually discard the information of the errors.
\comment{
\begin{figure}[!htbp]
    \centering
    \includegraphics[width=0.75\linewidth]{image_norm.png}
    \caption{Comparison between the input matrices $P_{\rho, i}$ (left) and its normalized version $\Bar{P}_{\rho, i}$ (right) [see Eqs. \eqref{Eq:DefImage} and \eqref{eq:threshold}].}
    \label{fig:before and after norm}
\end{figure}
}
In order to do this test, we slightly modify the images by normalizing the columns: $P_{\rho, i} \to \Bar{P}_{\rho, i}$ all columns of each image have the largest pixel equal to one. We also define an additional parameter \texttt{threshold} as follows:
\begin{align}
    &\text{If }\ \ \ \Bar{P}_{\rho, i} \equiv \frac{P_{\rho, i}}{\underset{i}{\text{max}}\ (P_{\rho, i})} < \texttt{threshold}, \nonumber \\
    &\text{Then, }\ \ \ \Bar{P}_{\rho, i} = 0 \label{eq:threshold}.
\end{align}

\begin{figure}[!htbp]
    \centering
    \begin{subfigure}%{0.5\textwidth}
      \centering
      \includegraphics[width=0.9\linewidth]{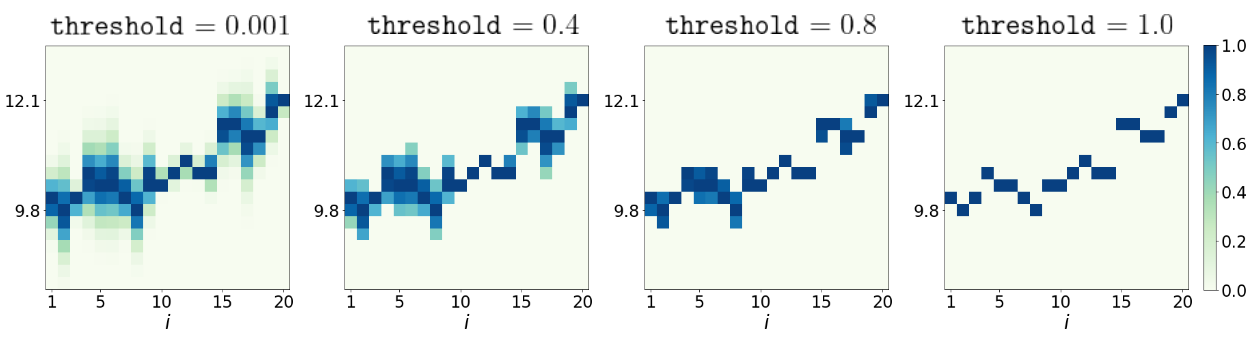}
      %\label{fig:su  %\caption{A subfigure}
    \end{subfigure}%
    \begin{subfigure}%{0.5\textwidth}
      \centering
      \includegraphics[width=0.7\linewidth]{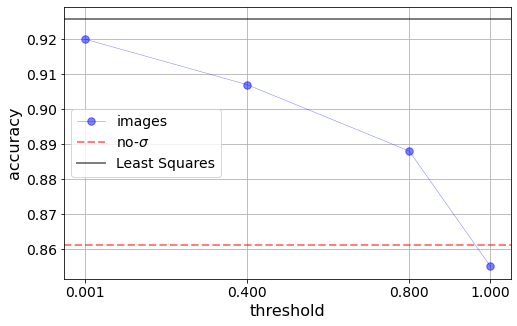}
      %\caption{A subfigure}
      %\label{fig:sub2}
    \end{subfigure}
    \caption{\textit{Top:} from left to right: input matrices of CNN images with \texttt{threshold} = $0.001, 0.4, 0.8$ and $1.0$ (see Eq. \eqref{eq:threshold}); \textit{Bottom:} accuracy as a function of \texttt{threshold}.}
    \label{fig:vary threshold}
\end{figure}

As we increase the \texttt{threshold}, the error bars in the image are “erased”, as shown in the top panel of Figure \ref{fig:vary threshold}. When \texttt{threshold} = 1, we are left only with the pixels representing the mean value. The bottom panel of Figure \ref{fig:vary threshold} shows how the accuracy changes as we vary this parameter. We see that the accuracy decreases as the error bars are discarded and, when the image contains only the mean values (\texttt{threshold} = 1), it approaches the accuracy obtained with CNN1D no-err.

\newpage

%{\noindent \em Remainder omitted in this sample. See http://www.jmlr.org/papers/ for full paper.}

\vskip 0.2in
\bibliography{sample}

\begin{thebibliography}{31}
\providecommand{\natexlab}[1]{#1}
\providecommand{\url}[1]{\texttt{#1}}
\expandafter\ifx\csname urlstyle\endcsname\relax
  \providecommand{\doi}[1]{doi: #1}\else
  \providecommand{\doi}{doi: \begingroup \urlstyle{rm}\Url}\fi

\bibitem[Abdar et~al.(2021)Abdar, Pourpanah, Hussain, Rezazadegan, Liu,
  Ghavamzadeh, Fieguth, Cao, Khosravi, Acharya, and et~al.]{Abdar_2021}
Moloud Abdar, Farhad Pourpanah, Sadiq Hussain, Dana Rezazadegan, Li~Liu,
  Mohammad Ghavamzadeh, Paul Fieguth, Xiaochun Cao, Abbas Khosravi, U.~Rajendra
  Acharya, and et~al.
\newblock A review of uncertainty quantification in deep learning: Techniques,
  applications and challenges.
\newblock \emph{Information Fusion}, 76:\penalty0 243–297, Dec 2021.
\newblock ISSN 1566-2535.
\newblock \doi{10.1016/j.inffus.2021.05.008}.
\newblock URL \url{http://dx.doi.org/10.1016/j.inffus.2021.05.008}.

\bibitem[Abu-Mostafa et~al.(2012)Abu-Mostafa, Magdon-Ismail, and
  Lin]{10.5555/2207825}
Yaser~S. Abu-Mostafa, Malik Magdon-Ismail, and Hsuan-Tien Lin.
\newblock \emph{Learning From Data}.
\newblock AMLBook, 2012.
\newblock ISBN 1600490069.

\bibitem[Acquarelli et~al.(2016)Acquarelli, van Laarhoven, Gerretzen, Tran,
  Buydens, and Marchiori]{article}
Jacopo Acquarelli, Twan van Laarhoven, Jan Gerretzen, Thanh Tran, Lutgarde
  Buydens, and Elena Marchiori.
\newblock Convolutional neural networks for vibrational spectroscopic data
  analysis.
\newblock \emph{Analytica Chimica Acta}, 954, 12 2016.
\newblock \doi{10.1016/j.aca.2016.12.010}.

\bibitem[Baldi et~al.(2014)Baldi, Sadowski, and Whiteson]{Baldi:2014kfa}
Pierre Baldi, Peter Sadowski, and Daniel Whiteson.
\newblock {Searching for Exotic Particles in High-Energy Physics with Deep
  Learning}.
\newblock \emph{Nature Commun.}, 5:\penalty0 4308, 2014.
\newblock \doi{10.1038/ncomms5308}.

\bibitem[Bi and Zhang(2005)]{NIPS2004_22b1f2e0}
Jinbo Bi and Tong Zhang.
\newblock Support vector classification with input data uncertainty.
\newblock In L.~Saul, Y.~Weiss, and L.~Bottou, editors, \emph{Advances in
  Neural Information Processing Systems}, volume~17. MIT Press, 2005.
\newblock URL
  \url{https://proceedings.neurips.cc/paper/2004/file/22b1f2e0983160db6f7bb9f62f4dbb39-Paper.pdf}.

\bibitem[Bousquet et~al.(2013)Bousquet, Boucheron, and Lugosi]{StatLearn}
Olivier Bousquet, Stéphane Boucheron, and Gábor Lugosi.
\newblock \emph{Introduction to Statistical Learning Theory}.
\newblock Springer, 2013.

\bibitem[Breiman et~al.(1984)Breiman, Friedman, Olshen, and
  Stone]{reason:BreFriOlsSto84a}
L.~Breiman, J.~H. Friedman, R.~A. Olshen, and C.~J. Stone.
\newblock \emph{Classification and Regression Trees}.
\newblock Wadsworth and Brooks, Monterey, CA, 1984.

\bibitem[Busca and Balland(2018)]{busca2018quasarnet}
Nicolas Busca and Christophe Balland.
\newblock Quasarnet: Human-level spectral classification and redshifting with
  deep neural networks, 2018.

\bibitem[Cabayol et~al.(2018)Cabayol, Sevilla-Noarbe, Fernández, Carretero,
  Eriksen, Serrano, Alarcón, Amara, Casas, Castander, and
  et~al.]{Cabayol_2018}
L~Cabayol, I~Sevilla-Noarbe, E~Fernández, J~Carretero, M~Eriksen, S~Serrano,
  A~Alarcón, A~Amara, R~Casas, F~J Castander, and et~al.
\newblock The pau survey: star–galaxy classification with multi narrow-band
  data.
\newblock \emph{Monthly Notices of the Royal Astronomical Society},
  483\penalty0 (1):\penalty0 529–539, Nov 2018.
\newblock ISSN 1365-2966.
\newblock \doi{10.1093/mnras/sty3129}.
\newblock URL \url{http://dx.doi.org/10.1093/mnras/sty3129}.

\bibitem[Caldeira and Nord(2020)]{Caldeira_2020}
João Caldeira and Brian Nord.
\newblock Deeply uncertain: comparing methods of uncertainty quantification in
  deep learning algorithms.
\newblock \emph{Machine Learning: Science and Technology}, 2\penalty0
  (1):\penalty0 015002, Dec 2020.
\newblock ISSN 2632-2153.
\newblock \doi{10.1088/2632-2153/aba6f3}.
\newblock URL \url{http://dx.doi.org/10.1088/2632-2153/aba6f3}.

\bibitem[Carleo et~al.(2019)Carleo, Cirac, Cranmer, Daudet, Schuld, Tishby,
  Vogt-Maranto, and Zdeborov\'a]{Carleo:2019ptp}
Giuseppe Carleo, Ignacio Cirac, Kyle Cranmer, Laurent Daudet, Maria Schuld,
  Naftali Tishby, Leslie Vogt-Maranto, and Lenka Zdeborov\'a.
\newblock {Machine learning and the physical sciences}.
\newblock \emph{Rev. Mod. Phys.}, 91\penalty0 (4):\penalty0 045002, 2019.
\newblock \doi{10.1103/RevModPhys.91.045002}.

\bibitem[Chang et~al.(2018)Chang, Cohen, and Ostdiek]{Chang:2017kvc}
Spencer Chang, Timothy Cohen, and Bryan Ostdiek.
\newblock {What is the Machine Learning?}
\newblock \emph{Phys. Rev. D}, 97\penalty0 (5):\penalty0 056009, 2018.
\newblock \doi{10.1103/PhysRevD.97.056009}.

\bibitem[Chollet et~al.(2015)]{chollet2015keras}
Fran\c{c}ois Chollet et~al.
\newblock Keras.
\newblock \url{https://keras.io}, 2015.

\bibitem[Czarnecki and Podolak(2013)]{inproceedings}
Wojciech Czarnecki and Igor Podolak.
\newblock Machine learning with known input data uncertainty measure.
\newblock volume 8104, pages 379--388, 09 2013.
\newblock \doi{10.1007/978-3-642-40925-7_35}.

\bibitem[Efron(1986)]{Efron:1986}
Bradley Efron.
\newblock {Why isn't everyone a Bayesian?}
\newblock \emph{Am. Stat.}, 40\penalty0 (1):\penalty0 1, 1986.

\bibitem[{Estrada} et~al.(2007){Estrada}, {Annis}, {Diehl}, {Hall}, {Las},
  {Lin}, {Makler}, {Merritt}, {Scarpine}, {Allam}, and
  {Tucker}]{2007ApJ...660.1176E}
J.~{Estrada}, J.~{Annis}, H.~T. {Diehl}, P.~B. {Hall}, T.~{Las}, H.~{Lin},
  M.~{Makler}, K.~W. {Merritt}, V.~{Scarpine}, S.~{Allam}, and D.~{Tucker}.
\newblock {A Systematic Search for High Surface Brightness Giant Arcs in a
  Sloan Digital Sky Survey Cluster Sample}.
\newblock \emph{Astroph. J.}, 660\penalty0 (2):\penalty0 1176--1185, May 2007.
\newblock \doi{10.1086/512599}.

\bibitem[{Firth} et~al.(2003){Firth}, {Lahav}, and
  {Somerville}]{2003MNRAS.339.1195F}
Andrew~E. {Firth}, Ofer {Lahav}, and Rachel~S. {Somerville}.
\newblock {Estimating photometric redshifts with artificial neural networks}.
\newblock \emph{Mon. Not. R. Astron. Soc.}, 339\penalty0 (4):\penalty0
  1195--1202, March 2003.
\newblock \doi{10.1046/j.1365-8711.2003.06271.x}.

\bibitem[Ismail~Fawaz et~al.(2019)Ismail~Fawaz, Forestier, Weber, Idoumghar,
  and Muller]{Ismail_Fawaz_2019}
Hassan Ismail~Fawaz, Germain Forestier, Jonathan Weber, Lhassane Idoumghar, and
  Pierre-Alain Muller.
\newblock Deep learning for time series classification: a review.
\newblock \emph{Data Mining and Knowledge Discovery}, 33\penalty0 (4):\penalty0
  917–963, Mar 2019.
\newblock ISSN 1573-756X.
\newblock \doi{10.1007/s10618-019-00619-1}.
\newblock URL \url{http://dx.doi.org/10.1007/s10618-019-00619-1}.

\bibitem[Kawamura et~al.(2021)Kawamura, Nishigaki, Andriamananjara,
  Rakotonindrina, Tsujimoto, Moritsuka, Rabenarivo, and
  Razafimbelo]{rs13081519}
Kensuke Kawamura, Tomohiro Nishigaki, Andry Andriamananjara, Hobimiarantsoa
  Rakotonindrina, Yasuhiro Tsujimoto, Naoki Moritsuka, Michel Rabenarivo, and
  Tantely Razafimbelo.
\newblock Using a one-dimensional convolutional neural network on visible and
  near-infrared spectroscopy to improve soil phosphorus prediction in
  madagascar.
\newblock \emph{Remote Sensing}, 13\penalty0 (8), 2021.
\newblock ISSN 2072-4292.
\newblock \doi{10.3390/rs13081519}.
\newblock URL \url{https://www.mdpi.com/2072-4292/13/8/1519}.

\bibitem[Kendall and Gal(2017)]{10.5555/3295222.3295309}
Alex Kendall and Yarin Gal.
\newblock What uncertainties do we need in bayesian deep learning for computer
  vision?
\newblock In \emph{Proceedings of the 31st International Conference on Neural
  Information Processing Systems}, NIPS'17, page 5580–5590, Red Hook, NY,
  USA, 2017. Curran Associates Inc.
\newblock ISBN 9781510860964.

\bibitem[Kingma and Ba(2017)]{kingma2017adam}
Diederik~P. Kingma and Jimmy Ba.
\newblock Adam: A method for stochastic optimization, 2017.

\bibitem[Le et~al.(2005)Le, Smola, and Canu]{10.1145/1102351.1102413}
Quoc~V. Le, Alex~J. Smola, and St\'{e}phane Canu.
\newblock Heteroscedastic gaussian process regression.
\newblock In \emph{Proceedings of the 22nd International Conference on Machine
  Learning}, ICML '05, page 489–496, New York, NY, USA, 2005. Association for
  Computing Machinery.
\newblock ISBN 1595931805.
\newblock \doi{10.1145/1102351.1102413}.
\newblock URL \url{https://doi.org/10.1145/1102351.1102413}.

\bibitem[Mehta et~al.(2019)Mehta, Bukov, Wang, Day, Richardson, Fisher, and
  Schwab]{MEHTA20191}
Pankaj Mehta, Marin Bukov, Ching-Hao Wang, Alexandre~G.R. Day, Clint
  Richardson, Charles~K. Fisher, and David~J. Schwab.
\newblock A high-bias, low-variance introduction to machine learning for
  physicists.
\newblock \emph{Physics Reports}, 810:\penalty0 1--124, 2019.
\newblock ISSN 0370-1573.
\newblock \doi{https://doi.org/10.1016/j.physrep.2019.03.001}.
\newblock URL
  \url{https://www.sciencedirect.com/science/article/pii/S0370157319300766}.
\newblock A high-bias, low-variance introduction to Machine Learning for
  physicists.

\bibitem[Mozaffari and Tay(2020)]{mozaffari2020review}
M.~Hamed Mozaffari and Li-Lin Tay.
\newblock A review of 1d convolutional neural networks toward unknown substance
  identification in portable raman spectrometer, 2020.

\bibitem[Murphy(2012)]{MurphyBook}
Kevin~P. Murphy.
\newblock \emph{The Machine Learning: A Probabilistic Perspective}.
\newblock MIT Press, 2012.

\bibitem[Nix and Weigend(1994)]{Nix1994EstimatingTM}
D.~Nix and A.~Weigend.
\newblock Estimating the mean and variance of the target probability
  distribution.
\newblock \emph{Proceedings of 1994 IEEE International Conference on Neural
  Networks (ICNN'94)}, 1:\penalty0 55--60 vol.1, 1994.

\bibitem[{Qu} et~al.(2021){Qu}, {Sako}, {M{\"o}ller}, and
  {Doux}]{2021arXiv210604370Q}
Helen {Qu}, Masao {Sako}, Anais {M{\"o}ller}, and Cyrille {Doux}.
\newblock {SCONE: Supernova Classification with a Convolutional Neural
  Network}.
\newblock \emph{arXiv e-prints}, art. arXiv:2106.04370, June 2021.

\bibitem[Reis et~al.(2018)Reis, Baron, and Shahaf]{Reis_2018}
Itamar Reis, Dalya Baron, and Sahar Shahaf.
\newblock Probabilistic random forest: A machine learning algorithm for noisy
  data sets.
\newblock \emph{The Astronomical Journal}, 157\penalty0 (1):\penalty0 16, Dec
  2018.
\newblock ISSN 1538-3881.
\newblock \doi{10.3847/1538-3881/aaf101}.
\newblock URL \url{http://dx.doi.org/10.3847/1538-3881/aaf101}.

\bibitem[{Storrie-Lombardi} et~al.(1992){Storrie-Lombardi}, {Lahav}, and
  {Sodre}]{1992MNRAS.259P...8S}
M.~C. {Storrie-Lombardi}, O.~{Lahav}, and Jr. {Sodre}, L.
\newblock {Morphological Classification of Galaxies by Artificial Neural
  Networks}.
\newblock \emph{Mon. Not. Roy. Astron. Soc.}, 259:\penalty0 8P, November 1992.
\newblock \doi{10.1093/mnras/259.1.8P}.

\bibitem[Tanabashi et~al.(2018)]{Tanabashi:2018oca}
M.~Tanabashi et~al.
\newblock {Review of Particle Physics}.
\newblock \emph{Phys. Rev. D}, 98\penalty0 (3):\penalty0 030001, 2018.
\newblock \doi{10.1103/PhysRevD.98.030001}.

\bibitem[Taylor(1997)]{TaylorBook}
John~R. Taylor.
\newblock \emph{An Introduction to Error Analysis: The Study of Uncertainties
  in Physical Measurements}.
\newblock University Science Books, 1997.

\end{thebibliography}

\end{document}